\pdfoutput=1

\documentclass[11pt,a4paper]{article}
\usepackage{CJKutf8}
\usepackage[dvipsnames]{xcolor}

\usepackage[acceptedWithA]{tacl2021v1}

\usepackage[margin=1in]{geometry}
\usepackage{tabularx}
\usepackage{tikz}
\usepackage{makecell}

\newcolumntype{M}{>{\centering\arraybackslash}m{1cm}}

\usepackage{times}
\usepackage{latexsym}
\usepackage[T1]{fontenc}

\usepackage[utf8]{inputenc}

\usepackage{microtype}

\usepackage[size=tiny]{todonotes}
\usepackage{tablefootnote}
\usepackage{enumitem}
\usepackage{amsmath}
\usepackage{booktabs}
\usepackage{graphicx}
\usepackage[normalem]{ulem}
\useunder{\uline}{\ul}{}
\usepackage{multirow}
\usepackage{subfigure}
\usepackage{float}
\usepackage{stfloats}
\usepackage{bbding}
\usepackage{tabularx}
\usepackage{colortbl} 
\usepackage{xcolor}
\usepackage{caption}

\definecolor{ref}{rgb}{0.0,1,0.94} 
\definecolor{r}{rgb}{1,1,0.4}

\newcommand{\yc}[1]{\textcolor{black}{#1}}

\newcommand{\seg}[1]{\textcolor{black}{#1}}
\newcommand{\serev}[1]{\textcolor{black}{#1}}
\newcommand{\sefinal}[1]{\textcolor{black}{#1}}
\newcommand{\ycrev}[1]{\textcolor{black}{#1}}
\newcommand{\yccr}[1]{\textcolor{black}{#1}}

\newcommand{\added}[1]{\textcolor{black}{#1}}

\newcommand{\cor}[1]{\textcolor{OliveGreen}{\textbf{#1}}}
\newcommand{\cors}[1]{\textcolor{OliveGreen}{#1}}

\newcommand{\incor}[1]{\textcolor{red}{\textbf{#1}}}
\newcommand{\incors}[1]{\textcolor{red}{#1}}

\newcommand{\hyppara}[0]{\textit{cand}$_{\textit{para}}$}
\newcommand{\hypadv}[0]{\textit{cand}$_{\textit{adv}}$}

\title{\includegraphics[scale=0.15]{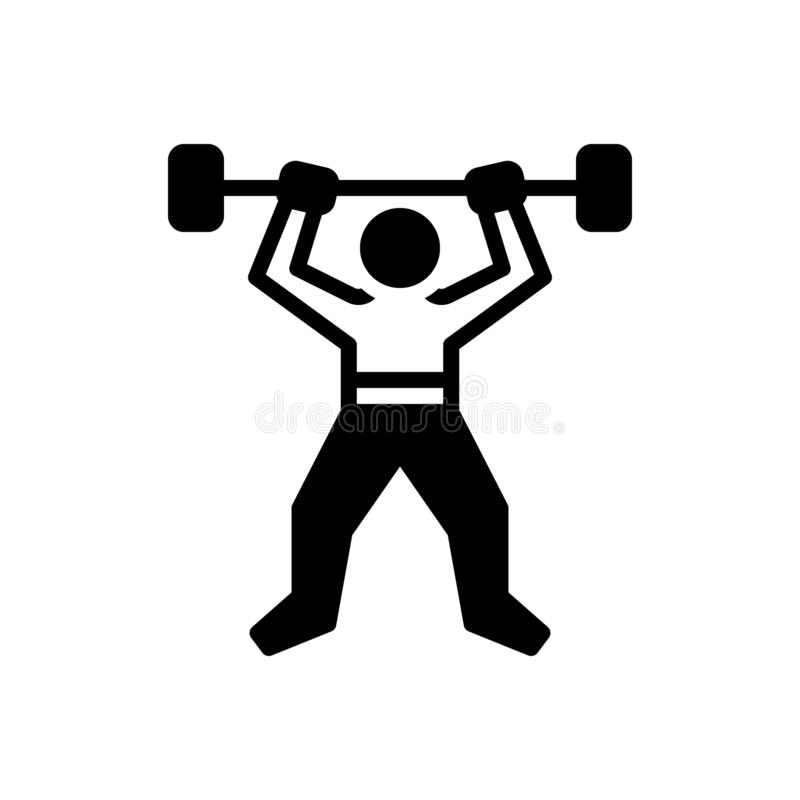}MENLI: Robust Evaluation Metrics from Natural Language Inference}

\author{
  \textbf{Yanran Chen}$^{1,2}$ \and \textbf{Steffen Eger}$^{2}$ \\
  $^{1}$Technische Universität Darmstadt, Germany \\
  $^{2}$\rm Natural Language Learning Group (NLLG), \url{https://nl2g.github.io/}\\
  Faculty of Technology, 
  Universität Bielefeld, Germany \\
  \texttt{yanran.chen@stud.tu-darmstadt.de, steffen.eger@uni-bielefeld.de} \\}

\begin{document}
\maketitle

\begin{abstract}
Recently proposed BERT-based evaluation metrics 
for text generation 
perform well on standard  
benchmarks but are vulnerable to adversarial attacks, e.g., relating to 
\added{information correctness}. We argue that this stems (in part) from  the fact that they are models of semantic similarity. In contrast, we develop evaluation metrics based on \emph{Natural Language Inference} (NLI), which we deem a more appropriate modeling. We design a preference-based adversarial attack framework and show that our NLI based metrics are much more robust to the attacks than the recent BERT-based metrics. On standard benchmarks, our NLI based metrics outperform existing summarization metrics, but perform below SOTA MT metrics. However, when combining existing metrics with our NLI metrics, we obtain both higher adversarial robustness (\added{15\%}–30\%) and higher quality metrics as measured on standard benchmarks (+5\% to \added{30\%}).
\end{abstract}

\section{Introduction}
Proper evaluation is key to fields such as machine learning and Natural Language Processing (NLP). Evaluation is particularly challenging for 
natural language generation (NLG) tasks, as there many be an infinitude of correct solutions (e.g., translations or summaries) for a given source text. While human evaluation is often considered the gold standard, it is slow and costly, thus researchers resort to automatic evaluation. 
Previously, this was done using simple lexical overlap metrics such as BLEU and ROUGE, but these exhibit low correlations with human judgments, 
particularly 
for 
state-of-the-art NLG 
systems 
\citep{mathur-etal-2020-tangled,peyrard-2019-studying}. 
Thus, a popular recent trend is to design automatic evaluation metrics based on 
large 
language models such as BERT  
and its many extensions \citep{bert-score,zhao-etal-2019-moverscore,sellam2020bleurt,wan-etal-2022-unite}. 

\begin{table}[!t]
    {\footnotesize
    \centering
    \setlength\tabcolsep{0.1pt}
    \noindent\begin{tabularx}{\columnwidth}{ >{\hsize=.35\hsize}X  >{\hsize=.65\hsize}X }
    
    \toprule
    \textbf{Concept} & \textbf{Examples} \\ \midrule
    Semantic Similarity & BERTScore, MoverScore, BaryScore, ...\\
    Text Generation & BARTScore, PRISM \citep{thompson-post-2020-automatic}\\
    Question Answering & QAEval \citep{deutsch-etal-2021-towards} \\
    NLI & MENLI\\
    \bottomrule
    \end{tabularx}%
    \caption{Different paradigms for metric induction proposed in recent years.} 
    \label{table:paradigm}
    }
\end{table}

Nonetheless, these novel metrics also have key limitations. For example, \citet{Sai_2021_EMNLP} and \citet{kaster-etal-2021-global} show that they are not robust to various adversarial attacks including lexical overlap and factuality errors. Taking the currently most popular metric---BERTScore\footnote{Published in 2020, BERTScore has more than 1700 citations as of March 2023.}---as an example, this adversarial vulnerability is unsurprising. BERTScore computes the semantic similarity between a reference and a 
system output 
(the \emph{candidate}), using a simplified token matching procedure. 
However, a good 
candidate 
is typically not appropriately identified by semantic similarity.  
For example, a 
candidate 
``5 Ukrainian soldiers wounded in Russia''
is 
not an adequate translation of a source corresponding to the reference ``50000 Russian soldiers killed in Ukraine'', 
although the two texts are of course semantically very similar.\footnote{That semantic similarity metrics are inherently incapable of identifying this puts them at great risk of being attacked by malicious agents, with serious real-world consequences, as the metrics cannot distinguish between truthful translations and semantically similar but factually incorrect translations.} 
While there have been many attempts to improve BERTScore using better token matching, e.g., using Word Mover Distance \citep{zhao-etal-2019-moverscore,chen-etal-2020-improving-text,colombo-etal-2021-automatic}, we argue that this line of research is a dead-end, as the underlying model of semantic similarity, originally proposed to address issues of lexical variation in BLEU/ROUGE, is simply not (fully) appropriate. 

An intuitively more suitable idea to model evaluation metrics is via \emph{natural language inference} (NLI) \citep{2013Dagan}. For example, in reference-based settings, \serev{in which 
{candidates} are compared to human references}, a candidate is intuitively good if it is \emph{equivalent} to a human reference via the concept of bi-implication. 
NLI systems are also promising alternatives because NLI is one of the most researched upstream tasks in NLP, where a lot of emphasis has been placed on concepts such as biases, generalization and adversarial conditions \citep{poliak-etal-2018-hypothesis,utama-etal-2020-mind}. 

In this paper, we  
ask whether we can directly use pre-trained NLI models as evaluation metrics, thereby establishing a new paradigm (but with predecessors, as indicated in  \S\ref{sec:related}). Our contributions: \begin{itemize}[topsep=5pt,itemsep=0pt,leftmargin=*]
    \item We design: a novel preference-based adversarial test suite for  
    MT and summarization metrics. 
    Our adversarial benchmark does not need human annotators, is suitable for \emph{reference-free} (\serev{where the candidate is directly compared to the source text, without human reference}) and \emph{reference-based} evaluation, and is challenging: e.g., BLEU, ROUGE, MoverScore and BERTScore perform below or at random level. 
        \item We explore: (i) how NLI metrics can be induced from existing NLI models; (ii) how they perform on benchmark and adversarial datasets, across (iii) two NLG problems, MT and summarization. 
    \item We show: (iv) NLI metrics perform particularly well in summarization, but below standard metrics in MT. 
    (v) They  
    substantially outperform existing metrics 
    on our adversarial attacks (e.g., 
    \added{$\sim$30\%–50\%} 
    margin 
    over the best \added{unsupervised} standard metric in MT).
    (vi) Combining existing metrics with our NLI metrics yields 
    both better (+5\%–30\%) and more robust metrics (+15\%–30\%). 
\end{itemize}

We point out that some 
current 
metrics already leverage NLI systems---thus, we do not include new information 
with respect to
them---but indirectly and thus (we argue) inadequately: e.g., MoverScore \citep{zhao-etal-2019-moverscore} leverages BERT representations  
fine-tuned on NLI. \citet{mathur-etal-2019-putting} train (pre-BERT) NLI-inspired architectures on MT datasets. In contrast, we show that by \emph{directly} leveraging NLI systems, much better adversarial and standard benchmark performances can be obtained. We call our novel metrics MENLI (\textbf{ME}trics from \textbf{NLI}).\footnote{Code$+$data: \scriptsize \url{http://github.com/cyr19/MENLI}.}

\section{Related Work}\label{sec:related}

Our work connects to evaluation metrics and NLI.

\paragraph{Evaluation Metrics for NLG}
In the last few years, researchers have come up with a plethora of different BERT-based metrics for varying tasks and setups: e.g., for MT and summarization, reference-based trained \citep{sellam2020bleurt,rei-etal-2020-comet} and untrained approaches \citep{zhao-etal-2019-moverscore,bert-score} have been suggested and the same is true for reference-free setups, where both supervised \citep{ranasinghe-etal-2020-transquest} and unsupervised metrics have been explored \citep{zhao-etal-2020-limitations,song2021sentsim,Belouadi2022USCOREAE}. In our work, we 
consider both reference-based 
as well as reference-free metrics.  
Both setups have important differences: Reference-free setups are more challenging, as they require to compare text in different languages (in MT) or of vastly different lengths (in summarization). On the other hand, they are more `resource-efficient', take humans out-of-the-loop, and promise web-scale evaluation. Both approaches are also different in terms of NLI. 
For example, while reference-based approaches require equivalence between reference and hypothesis, the concept of equivalence is not always appropriate in reference-free situations (e.g., in summarization, source and summary are intuitively not equivalent; rather,  source should entail summary). 

To realize metrics, different high-level 
approaches have been suggested as we outline in Table \ref{table:paradigm} (e.g., metrics from semantic similarity, from text generation or from question answering). There are also some predecessor works on metrics from NLI which we discuss below.

\paragraph{Robustness of Evaluation Metrics} has been a central issue of recent interest: \citet{Sai_2021_EMNLP} test metrics across several CheckList \citep{ribeiro-etal-2020-beyond} inspired templates, finding that most common standard metrics are not robust even to simple perturbations. \citet{kaster-etal-2021-global} probe metrics in an adversarial setting with lexical overlap, finding that they can be fooled by text that has high lexical overlap but low semantic similarity (indicating that the proposed BERT-based metrics are not even good models of semantic similarity). We combine the approaches of \citet{Sai_2021_EMNLP} and \citet{kaster-etal-2021-global}: While \citet{Sai_2021_EMNLP} use human crowd-workers to evaluate robustness, \citet{kaster-etal-2021-global} use a simpler preference-based setup, which does not need human annotators. We will also use the preference-based setup, 
but 
our attacks are largely inspired by \citet{Sai_2021_EMNLP}.

More recently (contemporaneously with us and after the first Arxiv submission of our work), several other papers have explored the robustness of recent evaluation metrics. 
\yccr{
For example, \citet{he2022blind} 
develop  stress test suites according to potential errors arising
from certain choices of metric design and pretrained language models 
used, 
showing that 
 metrics are biased towards their underlying models---e.g., 
BARTScore 
assigns higher scores to 
texts generated by the models of the metric itself.\footnote{ 
Robustness is also related to model \emph{biases}. For example, \citet{sun-etal-2022-bertscore} show that BERTScore encodes social biases such as gender biases. And \citet{deutsch-etal-2022-limitations} claim that reference-free metrics are inherently biased, which implies that they have unreasonable preferences. Our results show that many current reference-based metrics also have unreasonable preferences. Robustness checks are also related to \emph{explainability} \citep{Leiter2022TowardsEE,golovneva2023roscoe} of evaluation metrics as they help to understand metric limitations.}  
\citet{karpinska-etal-2022-demetr} explore the sensitivity of MT metrics to errors of different categories (regarding semantics, syntax and morphology) and severity, using a preference-based setting; they show that 
recent metrics like BERTScore 
dramatically outperform lexical overlap-based metrics such as BLEU and ROUGE, mostly obtaining over 95\% accuracy in their experiments.}
\yccr{
Our setups and that of \citet{karpinska-etal-2022-demetr} and \citet{he2022blind} are 
differentiated by the tasks considered, the preference specifications, the results, and the solutions proposed. \citet{karpinska-etal-2022-demetr} only evaluate metrics for MT while we consider both MT and summarization. They design their preferences in such a way that it would seem that recent metrics are 
quite robust while our more elaborate preferences 
expose their weak spots much better. 
Finally, we propose solutions (e.g., metrics from NLI) to addressing lack of robustness. 
Like us, \citet{he2022blind} also consider summarization and MT. Instead of designing preferences, however, they manually introspect how metric scores change as 
various perturbations are introduced. 
In this way, they expose blind spots of metrics. As remedies, they suggest to combine heterogeneous metrics to shield against varying blind spots (without performing concrete experiments)---we show that combining metrics with NLI based metrics yields additional robustness. 
}

Finally, 
\citet{rony-etal-2022-rome} develop RoMe as a robust metric in the context of semantic similarity, fluency and grammatical variability. 
They evaluate it on an adversarial dataset with five phenomena (entity, adjective and random word replacement; as well as text transformation and passive forms) by correlating against human judgments. 
Their model is a rather complicated 
trained metric leveraging semantic and grammatical features---we compare to it in  
\S\ref{sec:discussion}.

\paragraph{NLI}
NLI is one of the core upstream tasks in the NLP community. Due to its popularity, NLI has been investigated in-depth, where researchers found that trained models often overfit to low-level statistical cues instead of learning generalizable concepts of logical relationships between sentences \citep{poliak-etal-2018-hypothesis,gururangan-etal-2018-annotation}. As a consequence, many approaches to improve generalization have been investigated, e.g., \citep{belinkov-etal-2019-adversarial,utama-etal-2020-mind,zhou-bansal-2020-towards}. We argue that a high-quality NLI model would be an excellent candidate for an evaluation metric and explore this in this work.  

\sefinal{Like us, \citet{mathur-etal-2019-putting} note the similarity of (MT) evaluation and logical equivalence via NLI. They design supervised MT metrics leveraging different pre-BERT inspired architectures, including one from the NLI community called ESIM \citep{chen-etal-2017-enhanced} (which performs 
on par to an LSTM with attention in their experiments). 
Thus, in contrast to us, they do not leverage NLI models out-of-the-box as evaluation metrics but only fine-tune an NLI-inspired architecture on human scores from MT. 
}
MoverScore \citep{zhao-etal-2019-moverscore} fine-tunes BERT on NLI, which 
leads to better metrics. Thus, they, too, use NLI only indirectly. \citet{dusek-kasner-2020-evaluating} use NLI to evaluate hallucinations and omissions in reference-free data-to-text generation scenarios. They do not compare to any other metrics and do not consider NLI as a general paradigm for evaluation metrics. 
While the summarization community uses NLI models for \emph{consistency evaluation} \citep{fabbri2021summeval,laban-etal-2022-summac},  
to our knowledge, we are the first to verify the usefulness of NLI systems as \emph{general evaluation metrics} 
against a range of strong competitors, both in standard evaluation  
and  
adversarial attack settings. 

\section{Adversarial setup}
\label{sec:approach}

Following \citet{Sai_2021_EMNLP} and others, we consider an array of adversarial attacks on evaluation metrics---we will give a motivation of our attacks from the perspective of errors committed by real text generation systems below. 
In contrast to \citet{Sai_2021_EMNLP} \yccr{and similar to the later published work of \citet{karpinska-etal-2022-demetr}}, we implement a preference-based setup, which does not 
need human annotators. 
The advantages of the preference-based setup 
are: (i) lower cost (e.g., no annotation costs), (ii) which can be especially relevant for non-English languages (e.g., in ref-free situations for MT), and (iii) which allows adversarial evaluation at larger scale, yielding more robust estimates of performance. 
The challenge of the preference setup is to cleverly 
determine 
text pairs to compare. 

In our design, we 
use an anchor text (either the reference \emph{ref} or the source \emph{src}), a paraphrase 
\hyppara{} of the anchor text, and an adversarial text 
\hypadv{} which is maximally similar to the anchor text, but contains an adversarial attack. We 
expect a good metric $m$ to prefer 
\hyppara{} over 
\hypadv{}:
\begin{equation}
    \begin{split}\label{eq:pref}
\text{ref-based}: \:\: & m(\textit{ref}, \textit{cand}_{\textit{para}}) > m(\textit{ref}, \textit{cand}_{\textit{adv}})\\
\text{ref-free}: \:\: & m(\textit{src}, \textit{ref}) > m(\textit{src}, \textit{cand}_{\textit{adv}})
\end{split}
\end{equation}
 
The outcome of preferences in Eq.~(\ref{eq:pref}) depend on how we choose \hypadv{} and \hyppara{}, which we will describe below. In general, a challenging test suite has \hypadv{} maximally similar to \textit{ref}/\textit{src}, but with a key error. In contrast, \hyppara{} should be maximally dissimilar to \textit{ref}/\textit{src} (e.g., on surface level) but meaning-equivalent. 
Table \ref{tab:examples} illustrates the general structure of our adversarial test suite. 

\begin{table*}[!t]
\centering
\resizebox{1\textwidth}{!}{
\footnotesize
\noindent\begin{tabularx}{\textwidth}{ >{\hsize=.18\hsize}X | >{\hsize=.4\hsize}X | >{\hsize=.52\hsize}X }
\toprule
  &
  \textbf{Number error} &
  \textbf{Negation error} \\ \midrule
\textit{src} &
  Der bilaterale Handel wurde auf über \cors{100 Milliarden Dollar} im Jahr gesteigert. &
  Die Wirtschaft der Entwicklungs- und Schwellenländer \cors{wird schwach bleiben}. \\ \midrule

 \colorbox{ref}{\textit{ref}} &
  Bilateral trade has increased to more than \cors{\$100 billion} a year. &
  Emerging economies \cors{will remain weak}. \\ \midrule
\colorbox{r}{\textit{r}} (google translation of \textit{src}) &
  Bilateral trade has increased to over \cors{\$100 billion} a year. &
  The economies of developing and emerging countries \cors{will remain weak}. \\ \midrule
\hyppara{} &
  Bilateral trade has increased to more than \cors{one hundred billion dollars} a year. &
  Emerging markets \cors{will remain weak}. \\ \midrule 
\hypadv{}(\colorbox{ref}{ref-based}) &
  Bilateral trade has increased to more than \incors{\$814 billion} a year. &
  Emerging economies \incors{won't remain weak}. \\ \midrule
\hypadv{}(\colorbox{r}{ref-free}) &
  Bilateral trade has increased to over \incors{\$478 billion} a year. &
  The economies of developing and emerging countries \incors{won't remain weak}. \\ \bottomrule 

\end{tabularx}%
}
\caption{Examples of our adversarial test suite taken from WMT20$_{\text{de}}$. 
Red words indicate specific adversarial perturbations of the words in green.  
\hypadv{}(ref-based) builds on \textit{ref}, whereas \hypadv{}(ref-free) builds on \textit{r} 
(indicated by corresponding coloring in the first column). The preferences we query for are given in Eq.~\eqref{eq:pref}.
}
\label{tab:examples}
\vspace{-.4cm}
\end{table*}

\paragraph{\hypadv{}}

To obtain \hypadv{}, we consider the following attacks \added{(nine regarding information adequacy/correctness in 
\serev{candidates}
and three regarding text fluency)}, which we deem (to a large degree) representative for errors in different NLG tasks: 

\begin{itemize}[topsep=2pt,itemsep=-1pt,leftmargin=*]
    \item \emph{Addition}: 
    We randomly add a noun after an existing one 
    and
    connect them with ``and''. 
    For example, ``I love dogs'' 
    $\rightarrow$
    ``I love dogs and cats.''
    \item \emph{Omission}: 
    We use the framework 
    of \citet{Sai_2021_EMNLP} to 
    randomly drop 
    \ycrev{$\sim$1\%–20\% words in the sentence.} 
    \item \emph{Mismatch}: We  
    consider 
    mismatching nouns, verbs and adjectives, which can lead to misunderstanding of an entity, an action, and the speakers’ emotion, respectively. 
    Following \citet{chen-etal-2021-factuality-checkers}, 
    we replace a specific 
    word \ycrev{having the POS tag of noun/verb/adjective} 
    with another word having the same POS tag \ycrev{randomly selected from our collected words for that POS tag}. 
    \item \emph{Negation}: 
    We use the perturbation tool of 
    \citet{ribeiro-etal-2020-beyond} to add/remove negations to/from the verb for generating \hypadv{} 
    with contrary claims.
    \item \emph{Number error}: 
    We replace all numbers (except for those related to dates) in the sentence
    with random numbers in the same format (e.g., integer to integer, decimal to decimal). 
    \item \emph{Pronoun error}: 
    We replace \yccr{all pronouns in the sentence 
    with other ones} 
    without causing syntax errors 
    (e.g., ``he'' to ``she'' and ``us'' to ``them'').
   
    \item \emph{Name error}:  
    We use the tool of  
    \citet{ribeiro-etal-2020-beyond} to replace exactly one name 
    with a random one of the same gender. 
    \item \emph{Fluency}: We also include three phenomena from \citet{Sai_2021_EMNLP} to examine metrics' robustness against 
    attacks on  
    text fluency:
    (i) \emph{Jumbling word order}: Randomly shuffle the word order in a sentence.  
    \emph{Spelling error}: Add a typo to a word in a sentence.
    \emph{Subject-verb disagreement}: Make the subject and verb disagree  
    (e.g., ``He \emph{like} dogs.''). 
    
\end{itemize}

For ref-based metrics, we apply the perturbation templates 
 to \textit{ref} to construct \hypadv{}.   
 In contrast, for ref-free MT metrics, 
 we first translate 
 the source \textit{src} using 
 Google Translate 
 to a translation \textit{r} 
 and 
 then 
 perturb 
 \textit{r}  
 to obtain \hypadv{}. 
 We introduce $r$ to increase the similarity of \hypadv{} 
 to \textit{src}; e.g., we assume that Google Translate translates more literally, \serev{i.e., closer to word-by-word translations}, than 
 human translators. \serev{This may be important 
 to construct challenging test cases, 
 cf.\ \S\ref{sec:discussion} and our above discussion.} 
 \yc{For ref-free summarization, 
 we apply the perturbation templates to 
 a document \textit{r} which is maximally similar to 
 \textit{src}; 
 details follow.}

\begin{table}[!htb]
\footnotesize
\resizebox{\columnwidth}{!}{%
\noindent
\begin{tabularx}{\columnwidth}{ >{\hsize=.6\hsize}X | >{\hsize=.3\hsize}X | >{\hsize=.3\hsize}X | >{\hsize=.3\hsize}X | >{\hsize=.5\hsize}X | >{\hsize=.6\hsize}X}
\toprule
dataset            & task          & ref-based & ref-free &  \hyppara{} & \#examples \\ \midrule
PAWS$_{\text{ori}}$          & MT            & yes          & no         & ORI                                   & 2,000                                       \\
PAWS$_{\text{back}}$         & MT            & yes          & no         & BACK                                  & 2,000                                       \\
XPAWS$_{\text{x}}$
& MT            & yes          & yes        & ORI                                   & 455-474                                     \\
WMT20$_{\text{de}}$          & MT            & yes          & yes        & BACK                                  & 200                                         \\
SE$_{\text{adv}}$             & SUM & yes          & yes        & BACK                                  & 199                                         \\ \bottomrule
\end{tabularx}
}
\caption{Adversarial datasets. ``Yes/no'' indicates whether the dataset supports ref-based/free adversarial evaluation. ``ORI/BACK'' denotes whether \hyppara{} (except for \emph{number error}) 
is from the original datasets or backtranslation. ``\#examples'' refers to the avg.\ number of examples per phenomenon. XPAWS$_{\text{x}}$ denotes XPAWS$_{\text{de/fr/zh/ja}}$.}
\label{tab:adv_datasets}
\vspace{-.3cm}
\end{table}

\paragraph{\hyppara{}}
\serev{We use different ways to obtain \hyppara{}, because 
different kinds of paraphrases may yield more/less difficult test cases for metrics. We will analyze this in \S\ref{sec:discussion}. 
} 

In particular, we use 
data from
(1) PAWS \citep{paws2019naacl}, 
(2) PAWS-X \citep{pawsx2019emnlp}, 
(3) WMT20-news-commentary-v15 German-to-English 
\citep{mathur-etal-2020-results} to generate  \hyppara{} \yc{for MT evaluation metrics, and (4) SummEval for summarization metrics.} 
\serev{A summary with attributes is shown in Table \ref{tab:adv_datasets}.}

(1) PAWS contains sentence pairs created by word swapping and backtranslation, labeled as (non-)paraphrases by human raters. 
From  
sentence pairs 
labeled as \emph{paraphrase},  
we derive two datasets for ref-based evaluation metrics:
\begin{itemize}[topsep=2pt,itemsep=-1pt,leftmargin=*]
    \item \textbf{PAWS$_{\text{ori}}$}: We take the first sentence of a PAWS sentence pair as \textit{ref} and the second as \hyppara{}.
    \item \textbf{PAWS$_{\text{back}}$}: We use the first sentence of a PAWS sentence pair as \textit{ref} and  
    generate \hyppara{} based on \textit{ref} 
    using backtranslation (we use German as the pivot language)  
    except for {number error}, for which we replace the numbers in \textit{ref} with the corresponding words, using the Python library {num2words}.
\end{itemize}

(2) PAWS-X is the multilingual version of PAWS, which includes PAWS sentence pairs in six languages, translated from English PAWS,  allowing us to generate test suites for \sefinal{both} ref-free 
\sefinal{and ref-based metrics}. 
We use the first sentence in PAWS-X (e.g., German) 
as \textit{src} and 
the second sentence with the same ID in English PAWS as \textit{ref}. 
We select the data for two closer language pairs: German-to-English and French-to-English, and two more distant language pairs: Chinese-to-English and Japanese-to-English. Accordingly, we create 4 datasets: \textbf{XPAWS$_{\text{de}}$}, \textbf{XPAWS$_{\text{fr}}$}, \textbf{XPAWS$_{\text{zh}}$} and \textbf{XPAWS$_{\text{ja}}$}, each of which contains \textit{src} (first sentence of X-PAWS pair in source language), \textit{ref} (first sentence of English PAWS pair), 
and \hyppara{} (second sentence of English PAWS pair). 

(3) WMT20-news-commentary-v15 contains sentence pairs of source and human 
reference. 
From this, 
we create \textbf{WMT20$_{\text{de}}$}, directly taking the source and reference sentences as \textit{src} and \textit{ref}. We obtain \hyppara{} as in the case of \textbf{PAWS$_{\text{back}}$}. 

\yc{(4) SummEval \citep{fabbri2021summeval} contains documents and references from CNN DailyMail (CNNDM) 
\citep{hermann2015teaching}, 
with 10 additional human references. 
We rank the 11 references using ROUGE-L \citep{lin2004rouge} and 
use 
the reference $r$ with highest ROUGE score 
to generate \hypadv{} for ref-free setting, while the remaining 10 references serve as \textit{ref}.} We refer to \ycrev{the adversarial dataset induced from} SummEval as \textbf{SE$_{\text{adv}}$} in the remainder. 
We obtain \hyppara{} as in the case of \textbf{PAWS$_{\text{back}}$}.\footnote{
\yccr{As we generate our adversarial test instances fully automatically from backtranslation or automatic tools, 
they may contain some errors \sefinal{(including upper-/lower-case)}. 
For example, we note that in \hyppara{}, ``... billion dollars'' is sometimes incorrectly formulated as ``... dollars billion''; however, such cases occur only in $\sim$1\% of all test cases for number error, which we argue is still on an acceptable noise level. 
}}

\begin{table*}[!ht]
\centering

\resizebox{1\textwidth}{!}{
\footnotesize
\noindent\begin{tabularx}{\textwidth}{ >{\hsize=.18\hsize}X | >{\hsize=.36\hsize}X | >{\hsize=.56\hsize}X }
\toprule
  \textbf{Error} &
  \textbf{Source} &
  \textbf{MT hypothesis} \\ \midrule
\textit{Mismatch/verb} & \begin{CJK}{UTF8}{gbsn}\textcolor{OliveGreen}{关注}苏宁易购服务号\end{CJK}
   & 
  \textcolor{red}{Pay attention to} (\textcolor{OliveGreen}{Follow})  Suning.com service account \\ 
  \midrule
  \textit{Mismatch/adj.} & \begin{CJK}{UTF8}{gbsn}还不错，玩游戏的画质是真的\textcolor{OliveGreen}{香}\end{CJK} & 
  Not bad, the picture quality of playing games is really \textcolor{red}{fragrant} (\textcolor{OliveGreen}{good})
  \\ \midrule
  \textit{Pronoun/Addition} & \begin{CJK}{UTF8}{gbsn}买给儿子的，他说很好。\end{CJK} & Bought it for \textcolor{red}{his} (\textcolor{OliveGreen}{my}) son, he said it was good. \\\midrule
  \textit{Name} & \begin{CJK}{UTF8}{gbsn}当天，美国运输部长\textcolor{OliveGreen}{赵小兰}、美联邦众议员\textcolor{OliveGreen}{孟昭文}以及国际领袖基金会创会会长\textcolor{OliveGreen}{董继玲}等分别在会上发言。\end{CJK} & On the same day, US Secretary of Transportation \textcolor{red}{Zhao Xiaolan} (\textcolor{OliveGreen}{Elaine Lan Chao}), US Congressman \textcolor{red}{Meng Zhaowen} (\textcolor{OliveGreen}{Grace Meng}) and \textcolor{red}{Dong Jiling} (\textcolor{OliveGreen}{Chiling Tong}), founding president of the International Leaders Foundation, spoke at the meeting respectively. \\\midrule
  \emph{Omission} & I'll \textcolor{OliveGreen}{review} your account, one moment, please. & Ich werde Ihr Konto \textcolor{red}{[...]} (\textcolor{OliveGreen}{überprüfen}), einen Moment bitte.\\ \midrule
  \emph{Mismatch/noun} & Listen, I don't want to make \textcolor{OliveGreen}{my people} mad," she said. & „Hör zu, ich will \textcolor{red}{mein Volk} (\textcolor{OliveGreen}{meine Leute}) nicht verrückt machen“, sagte sie. \\ \midrule
  \emph{Pronoun} & Williams wasn't the only one who received a fine at this year's Wimbledon, though \textcolor{OliveGreen}{hers} was the most costly. & Williams war nicht die einzige, die beim diesjährigen Wimbledon eine Geldstrafe erhielt, obwohl \textcolor{red}{sie} (\textcolor{OliveGreen}{ihre}) die teuerste war. \\
 \bottomrule 

\end{tabularx}%
}
\caption{Examples of errors in WMT MQM annotations for Chinese-to-English and English-to-German. Red texts are the annotated errors (``[...]'' indicates the missing translation) and the green texts in the bracket refer to a more correct translation accordingly; the green texts in source sentences denote the parts being mistranslated or omitted.
}
\label{tab:real_errors_wmt}
\vspace{-.4cm}
\end{table*}

\vspace{-.4cm}
\yccr{
\paragraph{Real-world Motivation of Attacks}
Modern text generation systems are prone to many of the errors we investigate in this work. 
For example, 
\citet{freitag2021experts,freitag-etal-2021-results,freitag-etal-2022-results}
show, 
based on fine-grained human error annotations \citep{lommel-2014-mqm},  
that 
translations generated by state-of-the-art MT models still contain 
many accuracy-related errors (e.g., addition and omission of information, inappropriately informal pronouns) and sometimes even fluency-related errors (e.g., wrong spelling). 
Negation handling 
is also frequently discussed as an issue 
of modern MT
systems 
\citep{bentivogli-etal-2016-neural,sennrich-2017-grammatical,hossain-etal-2020-non,tang-etal-2021-revisiting}. 
In summarization, 
system summaries are often factually inconsistent with source documents in terms of numbers, named entities and assigning quotations to a particular person, etc. \citep{falke2019ranking,kryscinski-etal-2020-evaluating,chen-etal-2021-factuality-checkers}. 
More generally, \emph{hallucination} 
(of which addition/mismatches/etc.\ may be considered special cases) 
is a particular worrisome limitation of recent large language models   \citep{ji2022survey}. 
In Table \ref{tab:real_errors_wmt}, we show 
selected 
system translations from real MT systems with specific errors (following WMT MQM annotations) 
that are very similar to the ones we consider.\footnote{\url{https://github.com/google/wmt-mqm-human-evaluation}}
The frequency of errors 
may differ for various source-target language pairs (e.g., depending on their language distance) and formal/informal context. For example, when translating Chinese to English for news, the names are often directly translated to their Pinyin format (see the 4th row) 
instead of the official translations; in contrast, this rarely happens in English-to-German translations. 
But even for such closely related languages, 
NLG 
systems may omit information, or choose wrong pronouns, or mismatching nouns, particularly when a word has multiple senses. 
}

\section{Experimental Setup}
\subsection{Evaluation Metrics} 
\begin{table*}[!htb]
\captionsetup{justification=centering} 
\footnotesize
\resizebox{\textwidth}{!}{%
\noindent\begin{tabularx}{\textwidth}{ >{\hsize=.12\hsize}X  >{\hsize=.08\hsize}X | >{\hsize=.8\hsize}X }
\toprule
\multicolumn{2}{l|}{\textbf{Task}}                   & \textbf{Metrics}                                                      \\ \midrule
\multirow{2}{*}{MT}            & \text{ref-based} & MoverScore \citep{zhao-etal-2019-moverscore}, BERTScore \citep{bert-score}, BARTScore \citep{yuan2021bartscore}, SentSim \citep{song2021sentsim}, COMET \citep{rei-etal-2020-unbabels}, BLEURT \citep{sellam2020bleurt} \\
                               & ref-free  & COMET, SentSim, XMoverScore \citep{zhao-etal-2020-limitations}                                 \\ \midrule
\multirow{2}{*}{Summarization} & \text{ref-based} & BARTScore, DiscoScore \citep{zhao2022discoscore}, MoverScore, BERTScore            \\
                               & ref-free  & BARTScore, SUPERT \citep{gao2020supert}                                         \\ \bottomrule
\end{tabularx}
}
\caption{Evaluation metrics explored in this work.}
\label{tab:metrics}
\end{table*}
\ycrev{We explore a large array of recent state-of-the-art transformer based metrics, summarized in Table \ref{tab:metrics}. The variants used are briefly introduced below;
further details (e.g., model checkpoints and implementation) can be found on our Github.}

\ycrev{We report 
BERTScore F1 employing a RoBERTa-large model. For MoverScore, we use the unigram variant with a BERT-base model fine-tuned on MNLI \citep{N18-1101}. We use two variants of BARTScore (Precision and F1) for ref-based MT and summarization and BARTScore-FN (FN stands for Faithfulness) 
for 
ref-free summarization. 
We consider two variants of XMoverScore 
with different remapping strategies for multilingual embeddings (CLP, UMD) and two variants of SentSim 
with different word matching paradigms (BERTScore, WMD). We report the DiscoScore variant
with feature `Focus Frequency'.}

\subsection{Datasets \serev{\& Evaluation Protocol}}

\begin{table*}[]
\centering
\resizebox{\textwidth}{!}{%
\begin{tabular}{@{}cl|l@{}}
\toprule
\multicolumn{2}{l|}{\textbf{Task}}                                          & \multicolumn{1}{l}{\textbf{Datasets}}                                  \\ \midrule
\multirow{3}{*}{MT}                                & segment-level & WMT15-17, WMT20-21                          \\
                                                   & system-level  & WMT20-21                                                  \\
                                                   & adversary    & \emph{ref-based}: PAWS$_{\text{ori/back}}$, WMT20$_{\text{de}}$, XPAWS$_{\text{de}}$; 
                                 
\emph{ref-free}: XPAWS$_{\text{de/fr/zh/ja}}$, WMT20$_{\text{de}}$\\ \midrule
\multicolumn{1}{l}{\multirow{3}{*}{Summarization}} & summary-level & RealSum \citep{bhandari-etal-2020-evaluating}                                                       \\
\multicolumn{1}{l}{}                               & system-level  & RealSum, SummEval                                             \\
\multicolumn{1}{l}{}                               & adversary    & SE$_{\text{adv}}$, Rank19 \citep{falke2019ranking} (ref-free only)                                 \\ \bottomrule
\end{tabular}%
}
\caption{We use the to-English language pairs in WMT15-17 datasets \citep{stanojevic-etal-2015-results, bojar-etal-2016-results, bojar-etal-2017-results}. In segment-level evaluation on WMT20-21 \citep{mathur-etal-2020-results, freitag2021experts, freitag-etal-2021-results}, we use the data with MQM scores for zh-en, while in  system-level evaluation, we correlate the metrics with DA scores for all to-English language pairs. The datasets for system-level evaluation before WMT20 are skipped, as all metrics mostly get 
very high correlations on them.}
\label{tab:datasets}
\end{table*}

\ycrev{We summarize our used datasets in Table \ref{tab:datasets}. To evaluate the metrics' robustness under \textbf{adversarial conditions},
we use the datasets introduced in \S\ref{sec:approach}  
and additionally Rank19 \citep{falke2019ranking} (only for ref-free summarization), which contains examples composed of document paired with one correct and one incorrect candidate summary
with real-world factuality errors. 
In general, we check the metrics' preference between the two candidates and calculate \emph{accuracy}: 
the relative frequency  
that the metrics correctly choose among the two alternatives.
}

\ycrev{
On \textbf{MT} standard benchmarks, we evaluate the metrics on both 
\emph{segment-level} (where we correlate metrics scores to human judgements for individual sentences/segments in the datasets) and \emph{system-level}  
(where we correlate the average metric scores to the average human scores over the segments generated by each system),
using Pearson correlation 
as the performance indicator. 
On SummEval for \textbf{summarization}, we compute Kendall correlation with system-level human judgements on four criteria: \emph{coherence}, \emph{consistency}, \emph{fluency} and \emph{relevance} (we apply two aggregation methods for the multi-reference setting, \emph{max} and \emph{mean}).
We calculate Pearson correlation with both summary-level (analogous to segment-level in MT) 
and system-level \emph{LitePyramids} \citep{shapira2019crowdsourcing} human ratings in RealSumm. 
}

\subsection{NLI as a Metric}
NLI systems yield  
probability distributions over 
\emph{Entailment}, \emph{Contradiction} and \emph{Neutral}. 
We denote the probability values as \emph{e}, \emph{c}, and \emph{n}, where $e+c+n=1$ and $e,c,n\ge 0$. 
We first determine how to 
leverage the three values as NLI metrics. 

To do so, we evaluate \textbf{five simple formulas} of their arithmetic combination in a heuristic way: (1) \emph{e}, (2) \emph{-c}, (3) \emph{e-n}, (4) \emph{e-c} and (5) \emph{e-n-2c}, 
and inspect their effect in \textbf{three directions}, \serev{which correspond to the entailment directions implication, reverse implication and bi-implication}: 
    (i) \textbf{\emph{ref}/\emph{src} $\rightarrow$ \emph{cand}}, 
    where \emph{ref} or \emph{src} act as premise and \emph{cand} 
    as hypothesis; 
    (ii) \textbf{\emph{ref}/\emph{src} $\leftarrow$ \emph{cand}},  
    where \emph{cand} acts as premise and \emph{ref} or \emph{src} act as hypothesis and 
    (iii) \textbf{\emph{ref}/\emph{src} $\leftrightarrow$ \emph{cand}}, 
    as arithmetic average over the two above cases. 

For example, to obtain {\emph{e-n} from \emph{ref}/\emph{src} $\leftrightarrow$ \emph{cand}}, we first average the three probability scores over direction {\emph{ref}/\emph{src} $\rightarrow$ \emph{cand}} and {\emph{ref}/\emph{src} $\leftarrow$ \emph{cand}}, then
calculate \emph{e-n} based on the averaged scores. 
\yc{We only consider direction {\emph{src} $\rightarrow$ \emph{cand}} for ref-free summarization, since hypothesis does not need to entail source document.} 
The various selections of the formulas and directions result in 15 pooling strategies for NLI-based metrics. 

\paragraph{NLI Systems}
We explore both monolingual and cross-lingual
NLI-based metrics. 
For each setup, we choose two NLI models, which are obtained from Hugging Face or fine-tuning by ourselves. 

For \textbf{monolingual NLI metrics}, we choose (1) a RoBERTa-large model 
\citep{liu2019roberta} 
fine-tuned on SNLI \citep{bowman2015large}, MNLI, 
Fever \citep{nie2019combining} and ANLI 
\citep{nie-etal-2020-adversarial} 
by \citet{nie-etal-2020-adversarial} and (2) a DeBERTa-large model 
fine-tuned by \citet{he2021deberta}, using MNLI. We denote the NLI metrics induced from these two models
as \texttt{NLI-R} and \texttt{NLI-D}.
\yc{They will be used for ref-based MT evaluation, and both ref-based and -free summarization evaluation tasks.} 
Note that, while NLI-R has been fine-tuned on adversarial NLI (ANLI), which has been shown to increase robustness on (for example) negation and numerical reasoning, NLI-D has not been trained on ANLI. \textbf{Cross-lingual NLI metrics} 
should handle premises and hypotheses in different languages, so 
we select the multilingual versions of the underlying models of \texttt{NLI-R}/\texttt{NLI-D}. (1) We fine-tune a XLM-RoBERTa-base model \citep{xlm-roberta}, using the datasets for fine-tuning \texttt{NLI-R} as well as XNLI dataset \citep{conneau2018xnli}. (2) We select an mDeBERTa-base model 
fine-tuned on MNLI and XNLI. We denote the corresponding cross-lingual NLI metrics as \texttt{XNLI-R} and \texttt{XNLI-D}. 


\section{Experiment Results}\label{sec:results}

\serev{Before outlining our main results in \S\ref{sec:mt} (MT) and \S\ref{sec:summ} (summarization), we first discuss good pooling strategies for NLI metrics.}

\begin{table}[!htb]
\centering
\begin{subtable}[Reference-based]
{
\resizebox{\columnwidth}{!}{%
\begin{tabular}{@{}c|ccccc@{}}
\toprule
                             & e & -c & e-n & e-c & e-n-2c \\ \midrule
\emph{ref$\rightarrow$cand}            & 3+0 & 3+0  &    &  2+0  &       \\
\emph{ref$\leftarrow$cand}             &  &   &    &    &       \\
\emph{ref$\leftrightarrow$cand} & 0+4 &   & 0+3   & 0+1   & 0+2      \\ \bottomrule
\end{tabular}%
\centering
}
}
\end{subtable}

\begin{subtable}[Reference-free]
{
\resizebox{\columnwidth}{!}{%
\begin{tabular}{@{}c|ccccc@{}}
\toprule
                             & e & -c & e-n & e-c & e-n-2c \\ \midrule
\emph{src$\rightarrow$cand}            &  & 2+0  &    &    &       \\
\emph{src$\leftarrow$cand}               & 0+1 &   & 0+2   &    &       \\
\emph{src$\leftrightarrow$cand} & 0+1 &   & 4+6   & 4+0   &       \\ \bottomrule
\end{tabular}%
\centering
}
}
\end{subtable}
\caption{Winning frequency of different pooling strategies for NLI metrics on adversarial (first entry) and MT datasets (second entry). We only show non-zero entries.}
\vspace{-.4cm}
\label{tab:winning_ref_based}

\end{table}

\paragraph{Pooling Strategy}

\ycrev{We determine the 
pooling strategy for NLI metrics in \textbf{MT evaluation} from (1) the accuracy on the adversarial datasets and (2) the correlation with human judgements on the standard (segment-level) MT datasets. We leverage the \emph{winning frequency} of  
the pooling strategies to choose the best one; a strategy wins if it works best for an NLI metric among all 15 strategies. 
Overall, we find that the simple formula $e$ from the direction \emph{src/ref$\leftrightarrow$cand} is a good choice which works well for both standard and adversarial benchmarks, even though slightly better formulas could be chosen in selected subsettings (e.g., ref-based vs.\ ref-free evaluation), see Table \ref{tab:winning_ref_based} for examples. 
}

\ycrev{
For \textbf{summarization}, the situation is slightly more complex:
(1) {\emph{e-c} from direction \emph{ref$\leftarrow$cand}} performs best for {ref-based} NLI metrics; (2) {\emph{-c} from direction \emph{src$\rightarrow$cand}} is the best 
strategy for {ref-free} NLI metrics. Thus, we compare NLI metrics adopting these strategies with classic metrics.
}
\vspace{-.2cm}

\ycrev{
\serev{Even though we only looked at global aggregate statistics, we still observe that our method of identifying the pooling strategies above leveraged the data on which we will later evaluate the NLI metrics}.  
\serev{To avoid leaking information from the test set}, 
we evaluate NLI metrics on each dataset with the pooling strategy selected from the remaining datasets for that task in \S\ref{sec:discussion}. 
}

\subsection{Machine Translation}\label{sec:mt}
\begin{table}[!htb]
\footnotesize
\setlength\tabcolsep{1.9pt}
\resizebox{\columnwidth}{!}{%
\begin{tabular}{@{}lcccccccc@{}}
\toprule
\multicolumn{1}{c}{} & \multicolumn{4}{c|}{Adv.}                                      & \multicolumn{4}{c}{MT}                                        \\ \midrule
\multicolumn{1}{c}{} & \multicolumn{2}{c}{ref-based} & \multicolumn{2}{c|}{ref-free}  & \multicolumn{2}{c}{ref-based} & \multicolumn{2}{c}{ref-free}  \\
\multicolumn{1}{c}{} & all           & adeq.         & all           & \multicolumn{1}{c|}{adeq.}         & seg          & sys            & seg           & sys           \\ \midrule
\multicolumn{9}{l}{Supervised}                                                                                                                       \\ \midrule
COMET                & 67.4          & 67.0            & 76.8          & \multicolumn{1}{c|}{74.5}         & 0.676         & 0.808           & 0.620            & 0.698          \\
BLEURT               & 74.8          & 79.8          &      -         &     \multicolumn{1}{c|}{-}          & 0.708         & 0.807           &       -        &       -        \\ \midrule
\multicolumn{9}{l}{Unsupervised}                                                                                                                     \\ \midrule
sentBLEU             & 32.9          & 27.2          &      -         & \multicolumn{1}{c|}{-}              & 0.380           & 0.757           &      -         &    -           \\
Rouge                & 34.3          & 28.7          &       -        &  \multicolumn{1}{c|}{-}             & 0.425         & 0.774           &       -        &    -           \\
MoverScore           & 48.3          & 46.9          &         -      & \multicolumn{1}{c|}{-}              & 0.567         & \textbf{0.806}  &        -       &      -         \\
XMoverS(UMD)         &               &               & 74.5          & \multicolumn{1}{c|}{71.7}               & -            & -              & 0.400            & 0.672          \\
XMoverS(CLP)         &        -       &         -      & 73.8          & \multicolumn{1}{c|}{70.9}         & -            & -              & 0.422          & \textbf{0.673} \\
BERTS                & 65.3          & 60.9          &        -       & \multicolumn{1}{c|}{-}              & \textbf{0.620}  & 0.799           &       -        &        -       \\
BARTS-P              & 67.4          & 64.2          &         -      & \multicolumn{1}{c|}{-}              & 0.587         & 0.761           &               &               \\
BARTS-F              & 78.4          & 77.8          &          -     & \multicolumn{1}{c|}{-}              & 0.593         & 0.802           &         -      &        -       \\
SentS(BERTS)         & 68.1          & 67.8          & 62.7          & \multicolumn{1}{c|}{65.5}           & 0.612         & 0.401           & 0.421          & -0.021          \\
SentS(WMD)           & 62.1          & 61.9          & 63.0            & \multicolumn{1}{c|}{65.8}          & 0.607         & -              & \textbf{0.427} & -             \\ \midrule
\multicolumn{9}{l}{NLI-based}                                                                                                                        \\ \midrule
X(NLI)-R             & 85.0            & 92.1          & 70.5          & \multicolumn{1}{c|}{75.8}          & 0.451         & 0.756           & 0.221          & 0.335          \\
X(NLI)-D             & \textbf{86.6} & \textbf{92.3} & \textbf{79.3} & \multicolumn{1}{c|}{\textbf{85.8}} & 0.439         & 0.770             & 0.149          & 0.581          \\ \bottomrule
\end{tabular}
}
\vspace{-.2cm}
\caption{Pearson correlation  
with human judgments in WMT and accuracy (\%) on our adversarial datasets, averaged over datasets. The performance of ref-based COMET is averaged over WMT20$_{\text{de}}$ and XPAWS$_{\text{de}}$, since it also requires source texts as input. In bold: best results among all unsupervised metrics including the NLI-based metrics.}
\label{tab:mt_performance_v2}

\end{table}

\subsubsection{Adversarial Evaluation} 
\label{sec:adversarial_evaluation}

We now compare our NLI metrics with the best pooling strategy 
to 
our baseline metrics under adversarial conditions.

From Table \ref{tab:mt_performance_v2} \serev{(columns  ``Adv.'')}, 
we observe that in the \textbf{ref-based} setup: (1) NLI metrics outperform the great majority of metrics by a huge margin: over 85\% vs.\  32\%–78\% (all phenomena) and 92\% vs. 27\%–\ycrev{80}\% (\added{adequacy phenomena only}) 
on average. (2) Further, the two NLI metrics perform similarly. 
In the \textbf{ref-free} setup, 
the best cross-lingual NLI metric (\texttt{XNLI-D}) is still most robust under our attacks.
However, NLI metrics do not as substantially outperform the other metrics as in the 
ref-based setup. 
A potential reason is that 
the cross-lingual NLI models 
underperform compared to the monolingual setup (the preferences we query for in the reference-free setup may also play a role). Nevertheless, when excluding the fluency-related phenomena from the adversarial datasets,
\texttt{XNLI-D} is still on average 10 points better than the best standard metric, COMET (86\% vs.\ 75\%).

\begin{figure*}
\centering
    \includegraphics[scale=0.7]{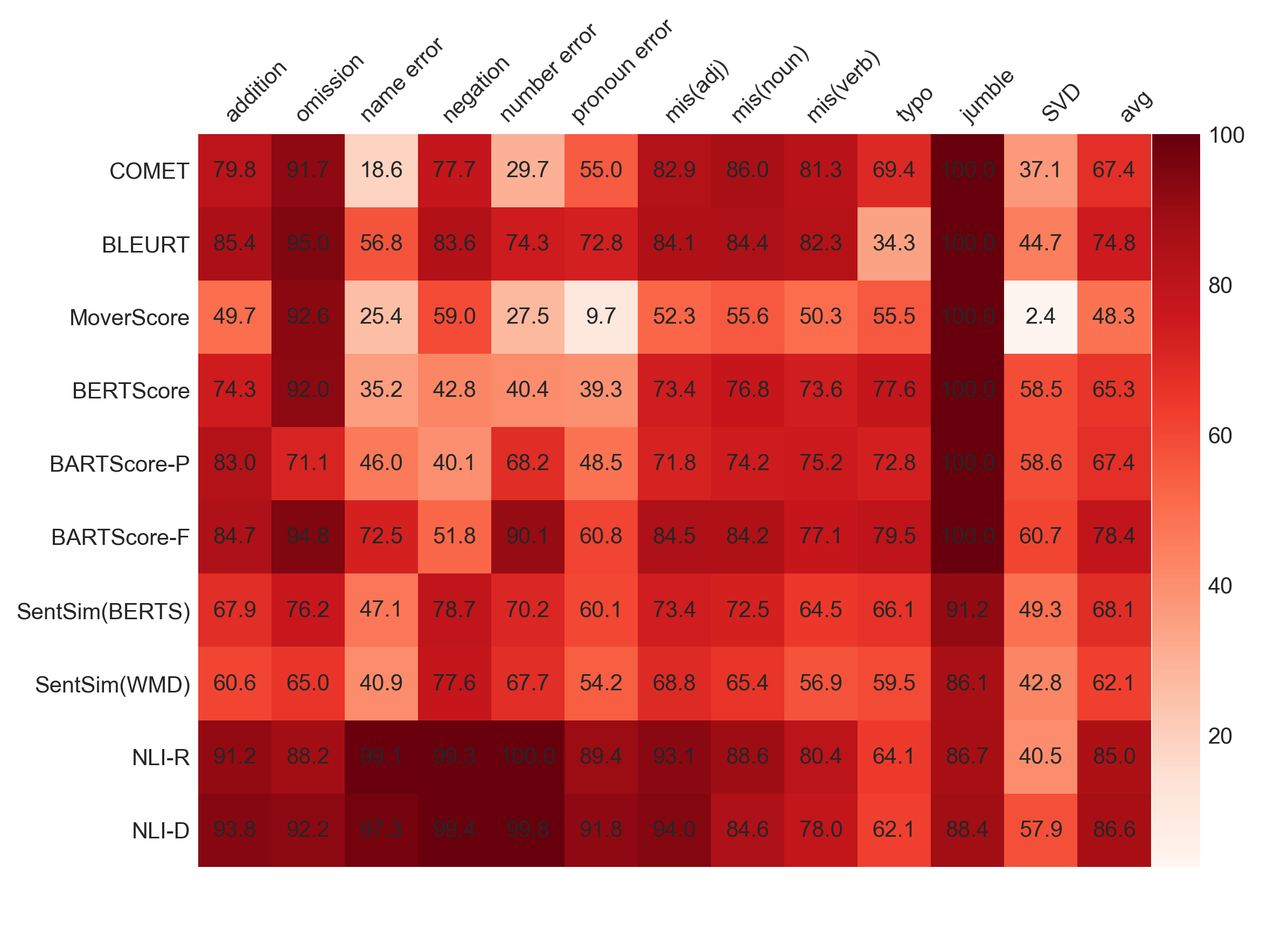}
    \caption{Average accuracy (values in each block) of all metrics per phenomenon over the adversarial datasets for ref-based MT evaluation. Darker color indicates higher accuracy and vice versa.}
    \label{fig:attack_matrix}
\end{figure*}

Moreover, our results 
reveal that: 
(1) most standard metrics are particularly incapable of detecting \emph{name error}, \emph{number error}, and \emph{pronoun error} ($\sim$29\%–70\%); (2) standard metrics, especially BLEURT and COMET, are most competitive regarding \emph{omission}, \emph{addition}, and 
\emph{jumbling} ($\sim$80\%–100\%); (3) NLI metrics are suboptimal for fluency 
attacks (mostly at random level), especially the reference-free NLI metrics 
and (4) 
NLI metrics 
are much better at 
\emph{name error}, \emph{negation}, 
\emph{number error}, \emph{pronoun error} and \emph{adj.\ mismatch} than most of the other metrics, especially ref-based (>90\% vs.\  
$\sim$10\%–80\%), as shown in Figure \ref{fig:attack_matrix}. 

\yccr{Our observations are inconsistent with \citet{karpinska-etal-2022-demetr}, where the state-of-the-art MT metrics mostly obtain >95\% accuracy in the preference-based evaluation. The reason is that our test suites are much more difficult for the evaluation metrics because we 
challenge 
them by lexical overlap between source/reference and candidate sentences during attacks: 
Metrics must choose between high lexical overlap adversarial candidates (with key errors) over low lexical overlap paraphrases. 
In contrast, in \citet{karpinska-etal-2022-demetr}, metrics are challenged to assign correct preferences for $\text{score}(\textit{ref},t)$ vs.\ $\text{score}(\textit{ref},t')$  where $t$ is a candidate and $t'$ the perturbed candidate. 
This 
is a much easier comparison because neither are $\textit{ref}$ and $t$ 
maximally dissimilar (but meaning equivalent) nor are 
$\textit{ref}$ and $t'$ maximally similar. This is an important lesson: \emph{How to design the adversarial preferences may critically affect the assessment of whether recent metrics are robust or not}. 
}

\subsubsection{Standard Benchmarks}
\label{sec:machine_translation}

\paragraph{Ref-based} We give average results over all datasets 
in Table \ref{tab:mt_performance_v2} 
(columns `MT'; individual results are available \sefinal{in our Github}). 
For \textbf{segment-level} evaluation, 
we 
observe: (1) trained metrics (COMET and BLEURT) substantially outperform the others, with average performance of 
\ycrev{$\sim$0.7 Pearson.}
(2) Unsupervised SOTA metrics have average correlation of 
\ycrev{$\sim$0.6 Pearson}, BERTScore is the best among them. 
(3) Our NLI-based metrics 
are not competitive, with correlations 
of 
\ycrev{$\sim$0.45 Pearson}.
When correlating with \textbf{system-level} human judgments, 
NLI metrics still underperform most of the SOTA metrics, 
but the margin is much smaller.

\paragraph{Ref-free} 
Trained metrics also 
dominate in \textbf{segment-level} 
evaluation ($>$\ycrev{0.6} Pearson), 
whereas the two NLI-based metrics perform much worse than the others 
\ycrev{(0.15-0.22 Pearson)}. Nevertheless, 
\texttt{XNLI-D} performs on par with COMET and better than the others 
on WMT20 at \textbf{system-level}. 

Overall, we conclude that 
our NLI metrics 
are not competitive with state-of-the-art evaluation metrics on standard MT datasets, especially at segment-level \seg{and ref-free}.

\subsubsection{Combined Metrics}
\label{sec:combine_mt}

\begin{figure}[!ht]
    \centering
    \includegraphics[width=\columnwidth]{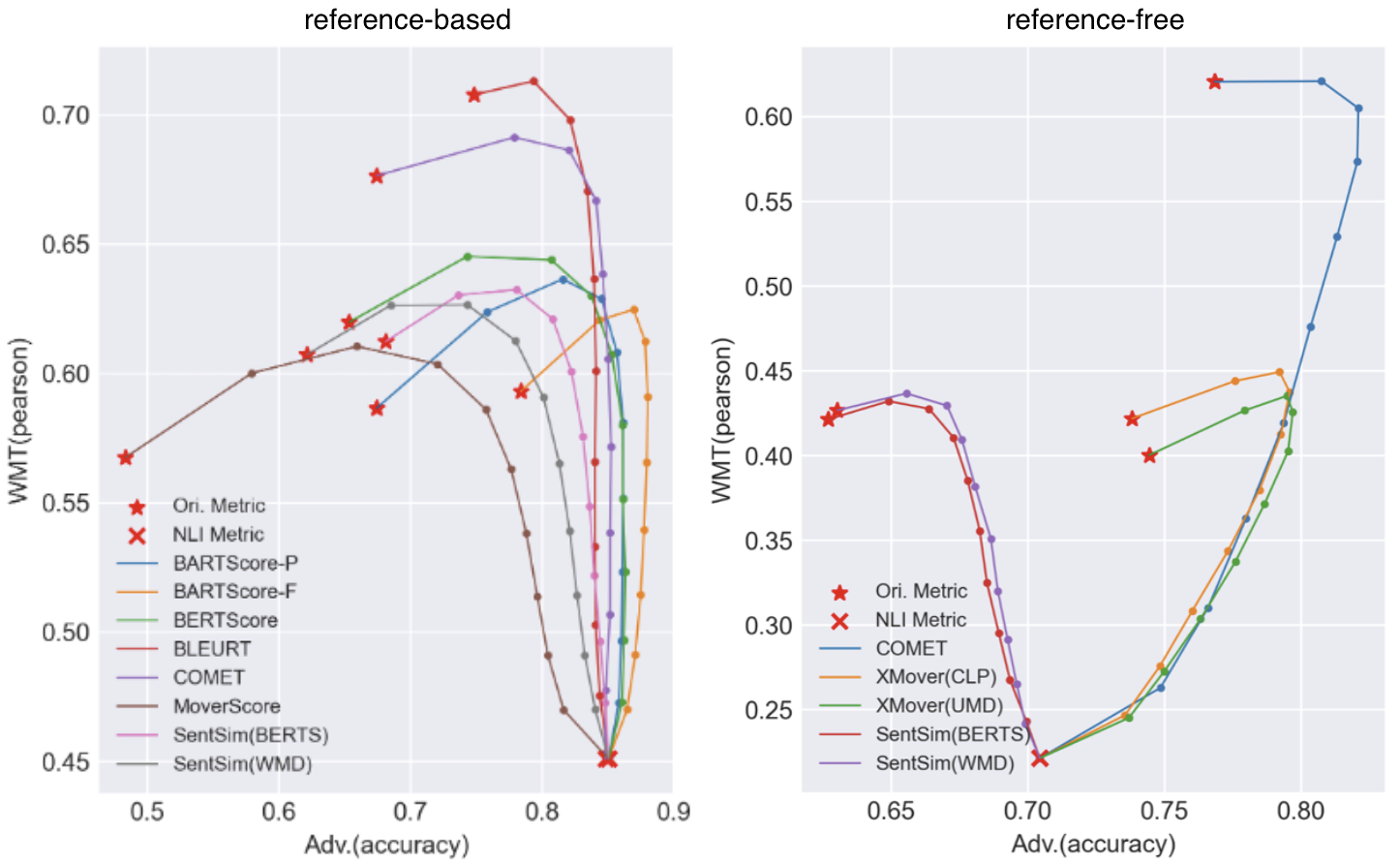}
    \caption{Accuracy on adversarial datasets and Pearson correlation with \textbf{segment-level} human judgements in WMT datasets of combined metrics with (X)NLI-R, averaged over datasets. 
    The points on each path from 
    the original metric to the 
    NLI metric indicate $w_{\text{nli}} = 0,0.1,. . .,1$.
    The purple line denoting the combination with ref-based COMET ends at another point since the corresponding adversarial performance is averaged over the 2 adversarial datasets containing source texts.}
    \label{fig:trade_off_R}
\vspace{-.4cm}
\end{figure}

Observing that NLI metrics are strong on adversarial setups, but comparatively weaker in standard evaluation, we examine \emph{how to get more robust metrics  which 
also perform well on 
standard benchmarks}.    
To do so, we take the weighted average of NLI and classical metrics: 
\begin{align}
    C &= w_{\text{nli}} \cdot N + (1-w_{\text{nli}}) \cdot M  
\end{align}
where 
$w_{\text{nli}} \in [0,1]$ is the weight for NLI metric $N$ 
and $M$ is a classical metric. 
Before combination, we rescale $M$ and $N$ to $[0, 1]$, using min-max normalization. 

We illustrate the performance of the combined evaluation metrics with \texttt{(X)NLI-R} 
on both adversarial and 
standard benchmarks \serev{(segment-level)} 
in Figure \ref{fig:trade_off_R}; the results for \texttt{(X)NLI-D} \serev{and for system-level} are similar. 
The x-axis denotes the average accuracy over the adversarial datasets, while y-axis is the average Pearson correlation over the
\sefinal{standard benchmarks (MT datasets).}
Each dot in each graph shows 
the value $C(w_\text{nli})$ for a specific weight $w_{\text{nli}}$. 
As seen from Figure \ref{fig:trade_off_R}, 
\seg{the graphs show an intriguing concave curvature.} 
In standard 
MT evaluation, 
the combination boosts the metric performance when $w_{\text{nli}}$ is small (from $0.1$ to $0.4$) in 
virtually all cases. 
We then see a \emph{simultaneous} increase of adversarial robustness and quality on standard benchmarks. In \textbf{ref-based} setup, e.g.\ for $w_{\text{nli}}=0.2$, we observe: (1) MoverScore and BARTScore-P improve most, 
with 
$\sim$8\%
\serev{(from 0.57/0.59 to 0.61/0.64 Pearson, respectively)} 
and 21\%–36\% improvements on adversarial datasets (from 48\%/\serev{67\%} to 66\%/\serev{82\%}
accuracy on average). 
(2) The best 
unsupervised metric on segment-level MT, BERTScore, 
increases
$\sim$4\% Pearson on standard benchmarks and 
$\sim$24\% accuracy 
on adversarial datasets. 
(3) 
The most robust untrained metric, BARTScore-F, improves about $\sim$11\% in robustness, whereas its performance on standard benchmarks also rises $\sim$5\%.
(4) The improvements on MT for trained metrics are smaller compared to those untrained metrics, with COMET improving only 1.5\% and BLEURT \sefinal{even becoming worse with the 
choice $w_{\text{nli}}=0.2$}.  
However, their performance in defending 
adversarial attacks still improves $\sim$10\%–20\%.
In \textbf{ref-free} setups,
all metrics improve $\sim$6\%–7\% on adversarial datasets.
Such setting only substantially boosts
XMoverScore's performance on standard 
benchmarks, with
$\sim$6\%–9\%.

\begin{figure}[!t]
    \centering
    \subfigure[MT]{
    \includegraphics[width=0.44\columnwidth]{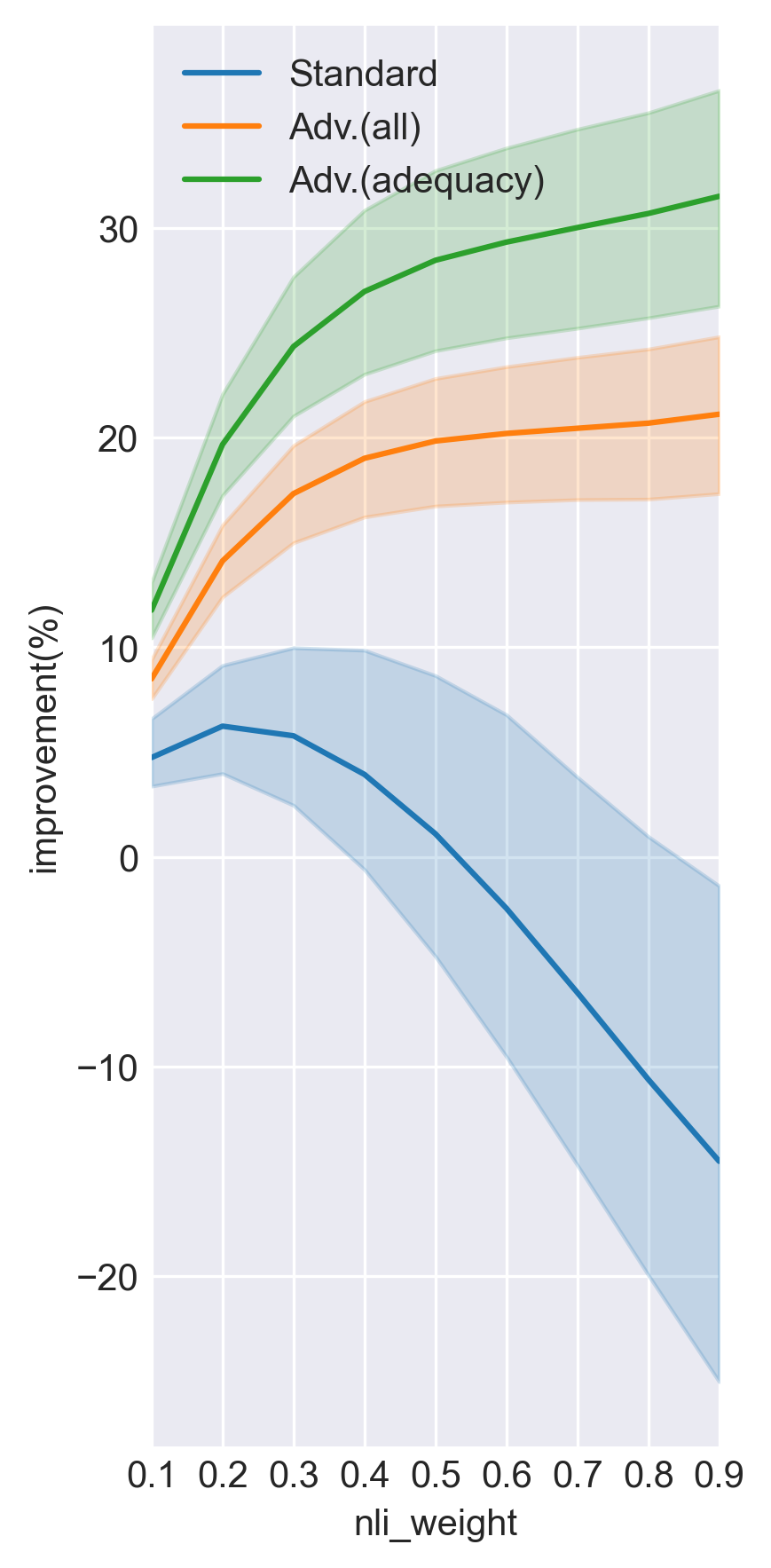}}
    \hspace{-2mm}
    \subfigure[Summarization]{
    \includegraphics[width=0.52\columnwidth]{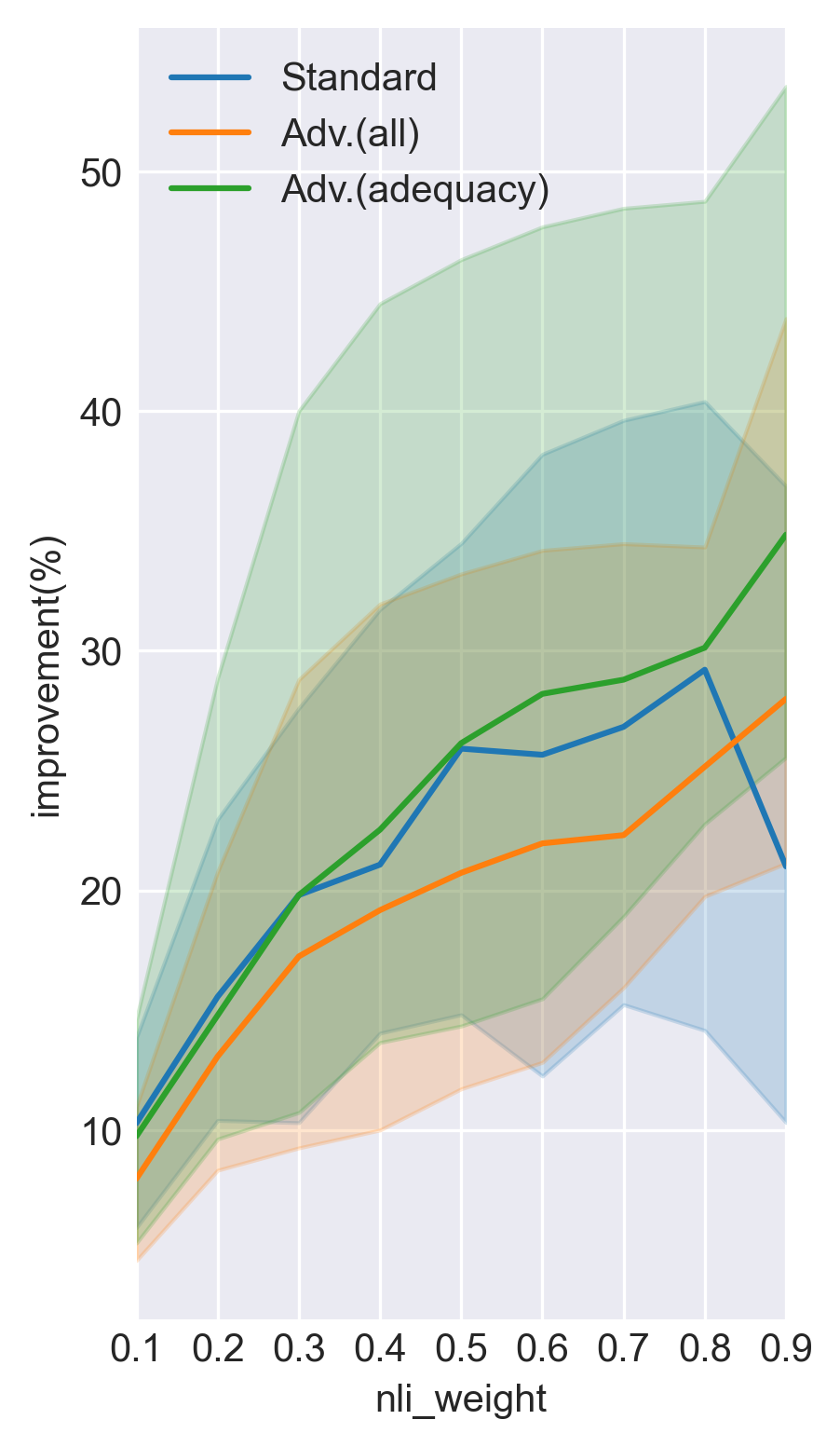}
    }
    \vspace{-0.3cm}
    \caption{Improvements of all metrics on standard benchmarks and adversarial datasets for $w_{\text{nli}}$ = 0.1,...0.9, averaged over all experiments. We show 95\% confidence interval.}
    \label{fig:confidence}
    \vspace{-.2cm}
\end{figure}

We summarize the improvements for all combinations in Figure \ref{fig:confidence}(a), which are averages over all experiments considered here. 
We can observe that the line denoting improvements on standard benchmarks peaks at $w_{\text{nli}}=0.2$, and the average improvements are positive when $w_{\text{nli}}\leq 0.5$. 
Further, on the adversarial datasets, 
the improvement monotonously increases with 
$w_{\text{nli}}$ 
and the gain is a concave function of $w_{\text{nli}}$ which saturates as $w_{\text{nli}}$ becomes larger. The sweet spots are $w_{\text{nli}} \in [0.2,0.3]$, which leads to 5\%–6\% improvement on standard benchmarks and 14\%–16\% improvement in adversarial robustness on average. 
When excluding the fluency phenomena 
from the adversarial datasets, the combined metrics consistently gain larger improvements in adversarial robustness, with 20\%-24\% improvements at the sweet spots.

\subsection{Summarization}\label{sec:summ}

\paragraph{Evaluation}

\begin{table*}[!htb]
\centering
\subfigure[Reference-based]{
\resizebox{.95\textwidth}{!}{%
\begin{tabular}{@{}lcccccccccccccc@{}}
\toprule
\multirow{3}{*}{metric} & \multicolumn{10}{c|}{SummEval}                                                                                                                                      & \multicolumn{2}{c|}{RealSumm}         & \multicolumn{2}{c}{Adv.}    \\ \cmidrule(l){2-15} 
                        & \multicolumn{2}{c}{coherence} & \multicolumn{2}{c}{consistency} & \multicolumn{2}{c}{fluency} & \multicolumn{2}{c}{relevance} & \multicolumn{2}{c|}{avg}            & \multicolumn{2}{c|}{litePyr}          & \multicolumn{2}{c}{SE$_{\text{adv}}$} \\
                        & mean          & max           & mean           & max            & mean         & max          & mean          & max           & mean  & \multicolumn{1}{c|}{max}    & sum & \multicolumn{1}{c|}{sys} & all           & adeq.        \\ \midrule
BLEU                    & 0.294         & 0.279         & 0.044          & -0.029         & 0.244        & 0.229        & 0.397         & 0.382         & 0.245 & \multicolumn{1}{c|}{0.215}  & 0.480   & \multicolumn{1}{c|}{0.124}  & 0.182         & 0.109        \\
Rouge                   & 0.191         & 0.176         & 0.088          & -0.279         & -0.037       & -0.081       & 0.118         & 0.103         & 0.090 & \multicolumn{1}{c|}{-0.020} & 0.540   & \multicolumn{1}{c|}{0.457}  & 0.185         & 0.117        \\ \midrule
MoverS              & 0.206         & 0.324         & 0.456          & 0.103          & 0.421        & 0.362        & 0.368         & 0.515         & 0.363 & \multicolumn{1}{c|}{0.326}  & \textbf{0.585}   & \multicolumn{1}{c|}{0.501}  & 0.287         & 0.251        \\
BERTS               & 0.618         & 0.618         & 0.221          & 0.044          & 0.273        & 0.185        & 0.603         & 0.515         & 0.429 & \multicolumn{1}{c|}{0.340}  & 0.574   & \multicolumn{1}{c|}{0.380}  & 0.598         & 0.574        \\
BARTS-P             & 0.485         & 0.441         & 0.176          & -0.044         & 0.376        & 0.185        & 0.500         & 0.368         & 0.385 & \multicolumn{1}{c|}{0.237}  & 0.478   & \multicolumn{1}{c|}{0.531}  & 0.697         & 0.692        \\
BARTS-F             & 0.515         & \textbf{0.647}         & 0.206          & 0.250          & 0.317        & 0.450        & 0.529         & \textbf{0.632}         & 0.392 & \multicolumn{1}{c|}{0.495}  & 0.583   & \multicolumn{1}{c|}{0.687}  & 0.788         & 0.792        \\
DiscoS              & \textbf{0.676}         & 0.279         & 0.279          & 0.676          & 0.539        & 0.554        & \textbf{0.632}         & 0.353         & \textbf{0.532} & \multicolumn{1}{c|}{0.466}  & -0.199  & \multicolumn{1}{c|}{-0.066} & 0.334         & 0.294        \\ \midrule
\multicolumn{15}{l}{NLI-based}                                                                                                                                                                                                                                       \\ \midrule
NLI-R                   & 0.147         & 0.074         & 0.632          & 0.676          & 0.494        & 0.450        & 0.279         & 0.206         & 0.388 & \multicolumn{1}{c|}{0.352}  & 0.525   & \multicolumn{1}{c|}{\textbf{0.856}}                       & \textbf{0.864}         & \textbf{0.905}        \\
NLI-D                   & 0.250         & 0.265         & \textbf{0.706}          & \textbf{0.750}          & \textbf{0.568}        & \textbf{0.613}        & 0.471         & 0.397         & 0.499 & \multicolumn{1}{c|}{\textbf{0.506}}  & 0.489   & \multicolumn{1}{c|}{0.840}  & 0.806         & 0.843        \\ \bottomrule
\end{tabular}
}
}
\subfigure[Reference-free]{
\resizebox{.95\textwidth}{!}{%
\begin{tabular}{@{}lccccccccccc@{}}
\toprule
\multirow{3}{*}{metric}   & \multicolumn{5}{c|}{SummEval}                                                                                                                                 & \multicolumn{2}{c|}{RealSumm}         & \multicolumn{4}{c}{Adv.}       \\ \cmidrule(l){2-12} 
                          & \multirow{2}{*}{coherence} & \multirow{2}{*}{consistency} & \multirow{2}{*}{fluency} & \multirow{2}{*}{relevance} & \multicolumn{1}{c|}{\multirow{2}{*}{avg}} & \multicolumn{2}{c|}{litePyr}          & \multicolumn{2}{c}{SE$_{\text{adv}}$} & Rank19 & \\
                          &                            &                              &                          &                            & \multicolumn{1}{c|}{}                     & summary & \multicolumn{1}{c|}{system} & all   & adeq. &  & avg   \\ \midrule
BARTS-FN & \textbf{0.735}                      & 0.132                        & 0.391                    & \textbf{0.662}                      & \multicolumn{1}{c|}{\textbf{0.480}}                & 0.178   & \multicolumn{1}{c|}{-0.023} & 0.427 & 0.389 & 0.796  & 0.612 \\
SUPERT           & 0.147                      & 0.603                        & \textbf{0.465}                    & 0.279                      & \multicolumn{1}{c|}{0.374}                & \textbf{0.522}   & \multicolumn{1}{c|}{0.626}  & 0.296 & 0.273 & 0.668  & 0.482 \\ \midrule
\multicolumn{12}{l}{NLI-based}                                                                                                                                                                                                                                     \\ \midrule
NLI-R                     & 0.221                      & 0.235                        & 0.391                    & 0.500                      & \multicolumn{1}{c|}{0.337}                & 0.300   & \multicolumn{1}{c|}{\textbf{0.688}}  & \textbf{0.720} & \textbf{0.722} & 0.866  & \textbf{0.793} \\
NLI-D                     & 0.162                      & \textbf{0.647}                        & 0.332                    & 0.324                      & \multicolumn{1}{c|}{0.366}                & -0.076  & \multicolumn{1}{c|}{0.568}  & 0.624 & 0.629 & \textbf{0.885}  & 0.755 \\ \bottomrule
\end{tabular}
}
}
\caption{Kendall correlation with system-level human judgments in SummEval. Pearson correlation with summary/system-level litePyramid in RealSumm. 
Accuracy on adversarial benchmarks, averaged over phenomena in SE$_{\text{adv}}$. We bold the best performance on each criterion. ``max/mean'' denotes the aggregation method used for multi-reference setting in ref-based evaluation on SummEval.}
\label{tab:sum_results}
\end{table*}

\ycrev{As Table \ref{tab:sum_results} shows, }
similar to MT evaluation, NLI-based metrics exhibit much stronger robustness under adversarial conditions (our best NLI metrics have at least 
\ycrev{$\sim$8}
points higher accuracy than the best standard metrics; \serev{right-most columns}). The difference is that the vanilla NLI metrics are now also comparably effective to the SOTA metrics 
on standard benchmarks. 
For instance, in \textbf{ref-based} setup, \texttt{NLI-D} with \emph{max} aggregation beats all metrics except for DiscoScore with \emph{mean}
on SummEval and both NLI metrics highly correlate with system-level human ratings in RealSumm (above 
\ycrev{0.8 Pearson}), where most standard metrics obtain only 
\ycrev{0.5–0.7 Pearson correlations}. 
When considering all evaluation dimensions of SummEval and RealSumm, NLI-D outperforms all other metrics, followed by NLI-R. Besides, we observe that NLI metrics correlate much better with human judgments regarding \emph{consistency} and (somewhat surprisingly) \emph{fluency} in SummEval compared to the other metrics. 
For the \textbf{ref-free} setup, 
BARTScore-FN performs best on SummEval---it outperforms the other metrics by above 
\ycrev{0.1 Kendall} on average. However, it does not correlate well
with both summary-level and system-level human judgments in RealSumm. NLI metrics 
are comparable or better than standard metrics on system-level. 
For example, \texttt{NLI-R} 
performs best among the examined metrics and is about 
\ycrev{0.06 Pearson} better than 
the best standard metric (SUPERT) 
on system-level 
in RealSumm. 
Nevertheless, reference-free NLI metrics also perform worse than the reference-based ones as in MT; an explicit bottleneck for the two NLI metrics is that they were
only trained on NLI data with short sentences, but reference-free summarization evaluation requires metrics 
\sefinal{to deal} with  
source documents which contain many more sentences. 

\vspace{-.2cm}
\paragraph{Combined Metrics}

In Figure \ref{fig:confidence}(b), we summarize the median improvements of combined summarization metrics (the median smooths some outliers).   
In contrast to MT, the combination brings 
almost equal benefits to performance of standard metrics on standard and adversarial benchmarks concerning only \added{adequacy}---we 
again observe a decrease in improvements on adversarial datasets when adding our fluency phenomena.
We 
identify a best $w_{\text{nli}}$, namely 0.8, with which the standard metrics gain about 25\%–30\% improvements in \emph{both} types of performances (adversarial and standard).

\begin{table*}[!ht]
\centering

\resizebox{1\textwidth}{!}{
\footnotesize
\noindent\begin{tabularx}{\textwidth}{ >{\hsize=.27\hsize}X | >{\hsize=.27\hsize}X | >{\hsize=.27\hsize}X | >{\hsize=.08\hsize}X | >{\hsize=.08\hsize}X | >{\hsize=.06\hsize}X}
\toprule
\textbf{\textit{ref}   }                                                                                                                                & \textbf{\hyppara   }                                                                                                         & \textbf{\hypadv }                                                                                                                              & \textbf{\textit{score}$_{\textit{para}}$: \textit{score}$_{\textit{adv}}$} (standard metric) & \textbf{\textit{score}$_{\textit{para}}$: \textit{score}$_{\textit{adv}}$} (NLI-R) & \textbf{error}   \\ \midrule
\multicolumn{6}{l}{BERTScore}                                                                                                                                                                                                                                                                                                                                                                                                                                                                           \\ \midrule

Although President George W. Bush says \textcolor{OliveGreen}{\textbf{he}} believes in markets, in this case \textcolor{OliveGreen}{\textbf{he}} has called for voluntary action.                         & Although President George W. Bush says \textcolor{OliveGreen}{\textbf{he}} believes in markets, \textcolor{OliveGreen}{\textbf{he}} has demanded voluntary action in this case.            & Although President George W. Bush says \textcolor{red}{\textbf{she}} believes in markets, in this case \textcolor{red}{\textbf{she}} has called for voluntary action.                       & 0.980: 0.982                              & 0.951: 0.000                         & \emph{Pronoun} \\ \midrule
\multicolumn{6}{l}{BARTScore-F}                                                                                                                                                                                                                                                                                                                                                                                                                                                                         \\ \midrule
\cor{Reagan} and I were nonetheless able to create a reservoir of constructive spirit through constant outreach and face-to-face interaction. & Nevertheless, \cor{Reagan} and I were able to create a constructive climate through constant contact and personal interaction. & \incor{Nicole} and I were nonetheless able to create a reservoir of constructive spirit through constant outreach and face-to-face interaction. & \makecell[l]{-2.104: \\-1.527}                            & 0.943: 0.002                         & \emph{Name}    \\ \midrule

\multicolumn{6}{l}{BLEURT}                                                                                                                                                                                                                                                                                                                                                                                                                                                                              \\ \midrule
In 2012, when Freedom House downgraded \cor{Mali} to “not free,” engagement declined by 7\%.                                                  & In 2012, when Freedom House classified \cor{Mali} as unfree, the engagement fell by 7 percent.                                 & In 2012, when Freedom House downgraded \incor{Melissa} to “not free,” engagement declined by 7\%.                                               & 0.787: 0.834                              & 0.983: 0.030                         & \emph{Name}    \\\midrule
This leads to heavy deforestation and lethal indoor air pollution, which kills \cor{1.3 million} people each year.                            & This leads to heavy Deforestation and lethal indoor air pollution, which kills \cor{one point three million} people each year. & This leads to heavy Deforestation and lethal indoor air pollution, which kills \incor{6.9 million} people each year.                            & 0.682: 0.767                              & 0.783: 0.000                         & \emph{Num}     \\\midrule
\multicolumn{6}{l}{COMET}              \\ \midrule

Who serves as president of the United States \cor{is} critically important for Mexicans. & Anyone who serves as President of the United States \cor{is} crucial to Mexicans. & Who serves as president of the United States \incor{is not} critically important for Mexicans. & 1.067: 1.086 & 0.974: 0.044 & \emph{Negation}
\\ \bottomrule

\end{tabularx}%
}
\caption{Sample instances in adversarial datasets where standard metrics failed while NLI-R succeeded; ref-based setup. In the 4th and 5th columns, we show \emph{[score assigned to \hyppara]: [score assigned to \hypadv]} by standard metrics and NLI-R, respectively; robust metrics should give \hyppara{} higher scores. Green bold texts indicate the anchor words/phrases to be perturbed and the red ones in \hypadv{} refer to the corresponding perturbed texts.}
\label{table:failure_examples}
\vspace{-.4cm}
\end{table*}

\vspace{-0.1cm}
\section{Discussion \& Analysis}\label{sec:discussion}
\vspace{-.1cm}

\paragraph{Selected Failure Cases of Metrics:} Table \ref{table:failure_examples} shows selected failure cases of four popular 
metrics (BERTScore, BARTScore, BLEURT, COMET), 
where the \sefinal{NLI} metrics are correct in each case. In the examples, BERTScore 
prefers text with the wrong gendered pronoun over a legitimate paraphrase and even trained metrics like BLEURT fail on severe name changes such as ``Melissa'' (a person name) vs.\ ``Mali'' (a country name). Leveraging more 
subtle cases (e.g., mismatches based on wrong word senses instead of random mismatches with the same POS or replacing names with names of the same `type') would 
likely constitute even harder test cases for future metrics. 

\paragraph{No Metric is Good Everywhere:} Across distinct dimensions, different metrics perform differently, indicating that they capture varying aspects. 
For example, NLI metrics are not so good on fluency adversarial attacks, e.g., typos. This may be unsurprising, given that fluency is a low-level phenomenon while NLI concerns high-level logical relationships between sentences (some fluency phenomena would best be treated by switching to a lower-level representation space, such as character-level \citep{Vu2022LayerOR}; this could seamlessly be integrated in existing NLI models).  
The NLI metrics are also weaker concerning segment-level MT evaluation on standard benchmarks. However, NLI metrics alone perform surprisingly well: 
In ref-based MT, they win on 7 out of 19 dimensions (12 adversarial phenomena and 7 standard datasets, evaluated segment- and system-level), only beaten by BLEURT (8 wins); ref-free,  they win 5 out of 19 dimensions, second only to COMET (11 wins). In ref-based summarization, they are clearly ahead of all standard metrics, winning not only 8 out of 12 adversarial dimensions, but also system-level LitePyramid, consistency and fluency (thus, 11 out of 18 wins), clearly ahead of BARTScore-P (4 of 18); ref-free, they are also best 
and 
win 13 out of 18 dimensions. The best overall metrics, measured as average performance over standard and adversarial datasets, always include NLI: for ref-based MT, this is BLEURT+$0.2\times$NLI-R, for ref-free MT, it is COMET+$0.3\times$NLI-D. For summarization, NLI-R alone and combined with BARTScore-F perform best on average. 

\vspace{-.1cm}
\paragraph{Rescaling:} The min-max normalization 
we used  (a standard technique for normalizing data in machine learning, typically applied to input features) 
for metric combination
requires batch processing.
It is necessary to account for the different ranges of metrics, e.g., some metrics take negative values. An alternative would include to enforce more formal constraints on evaluation metrics, i.e., that they should take outputs in [0,1]. 
When applying our combined metrics in practice, one could also replace them by surrogate metrics trained on the outputs of the original combined metrics or simply take the min-max values inferred from the datasets already evaluated on---the larger these datasets the more reliably are min and max estimated.  

\paragraph{Sensitvity to $w_{\text{nli}}$:} 
Having different weights $w_{\text{nli}}$ for different tasks is undesirable, because it requires 
considering each task individually. However, in our experiments, we found that all small $w_{\text{nli}}$ (below $0.5$) yield good performances and are thus safe choices: They increase adversarial robustness and also lead to better metrics 
on standard benchmarks. 

\paragraph{Adversarial Performance vs.\ Standard Performance:} 
\ycrev{
From our experiments, it might seem that adversarial and standard performance are anti-correlated: A metric with higher adversarial performance may have lower performance on standard benchmarks and vice versa. While this would not necessarily be a major surprise as adversarial conditions oftentimes test phenomena that are otherwise not represented in standard benchmarks \citep{niven-kao-2019-probing}, 
a statistical analysis reveals 
that standard performance generally \emph{positively} correlates to the adversarial performance in our case, \yccr{consistent with our earlier argument that existing NLG systems in the real world do commit similar errors as we check for}. 
To do so, we first convert the metrics’ standard performance to rankings for each performance category (e.g., ref-based/-free segment/system-level MT performance,  performance on SummEval/RealSumm), then we correlate the ranking-based standard performance 
to the corresponding adversarial performance rankings, obtaining 0.37 Spearman. When excluding NLI metrics, the correlation increases to 0.60.
}

\paragraph{The Choice of \hyppara{} Matters:}
\ycrev{As indicated in \S\ref{sec:approach}, we speculate that a good adversarial setting maximizes (surface) dissimilarity between \emph{ref} and \hyppara{} (which can better trick the metrics). To investigate, we compute the normalized 
edit distance between \emph{ref} and \hyppara{};\footnote{Ref-free, the edit distance between \emph{r} and \emph{ref} is considered.} a larger edit distance means a greater dissimilarity. 
If our assumption is true, then 
larger edit distances 
represent 
harder test cases for the metrics.}
\ycrev{
We find: 
(1) the average edit distance for the test cases where the metrics fail to defend against the adversarial attacks is 
0.01–0.6 
larger than that for where they succeed, averaged over metrics; 
(2) 
for PAWS$_{\text{back}}$ and PAWS$_{\text{ori}}$ (both induced from PAWS) where the \hyppara{} are obtained in different ways, all metrics achieve 0.02–0.15 lower accuracy on PAWS$_{\text{ori}}$, which has 
0.46 
larger average edit distance than PAWS$_{\text{back}}$, in turn. Both findings confirm our above assumption. 
In addition, we observe that NLI metrics have the smallest difference between the edit distances for failure and success cases 
(0.01–0.26)
as well as that between the accuracy on PAWS$_{\text{back}}$ and PAWS$_{\text{ori}}$ (0.02) among all evaluation metrics.  
This implies that they are least affected by 
surface overlap 
and instead better consider the logical relationship between sentences. This is what makes them attractive as evaluation metrics. 
}

\paragraph{The Choice of \hypadv{} Matters, Too:}
\ycrev{
We evaluate on one complex attack combining \emph{Number error} with \emph{Negation} which increases the difference between \textit{ref} and \hypadv{} based on the test cases for \emph{Number error} in WMT20$_{\text{de}}$. The accuracy increases by an average of 0.28 over all metrics.
This confirms our assumption that maximizing the (surface) similarity between \textit{ref} and \hypadv{} (but with key errors) leads to harder test suites and vice versa. 
}

\vspace{-.4cm}
\paragraph{Ensemble with NLI Metrics Are More Effective:}\label{sec:ensemble_other}

\begin{figure}
    \centering
    
    \includegraphics[width=.65\columnwidth]{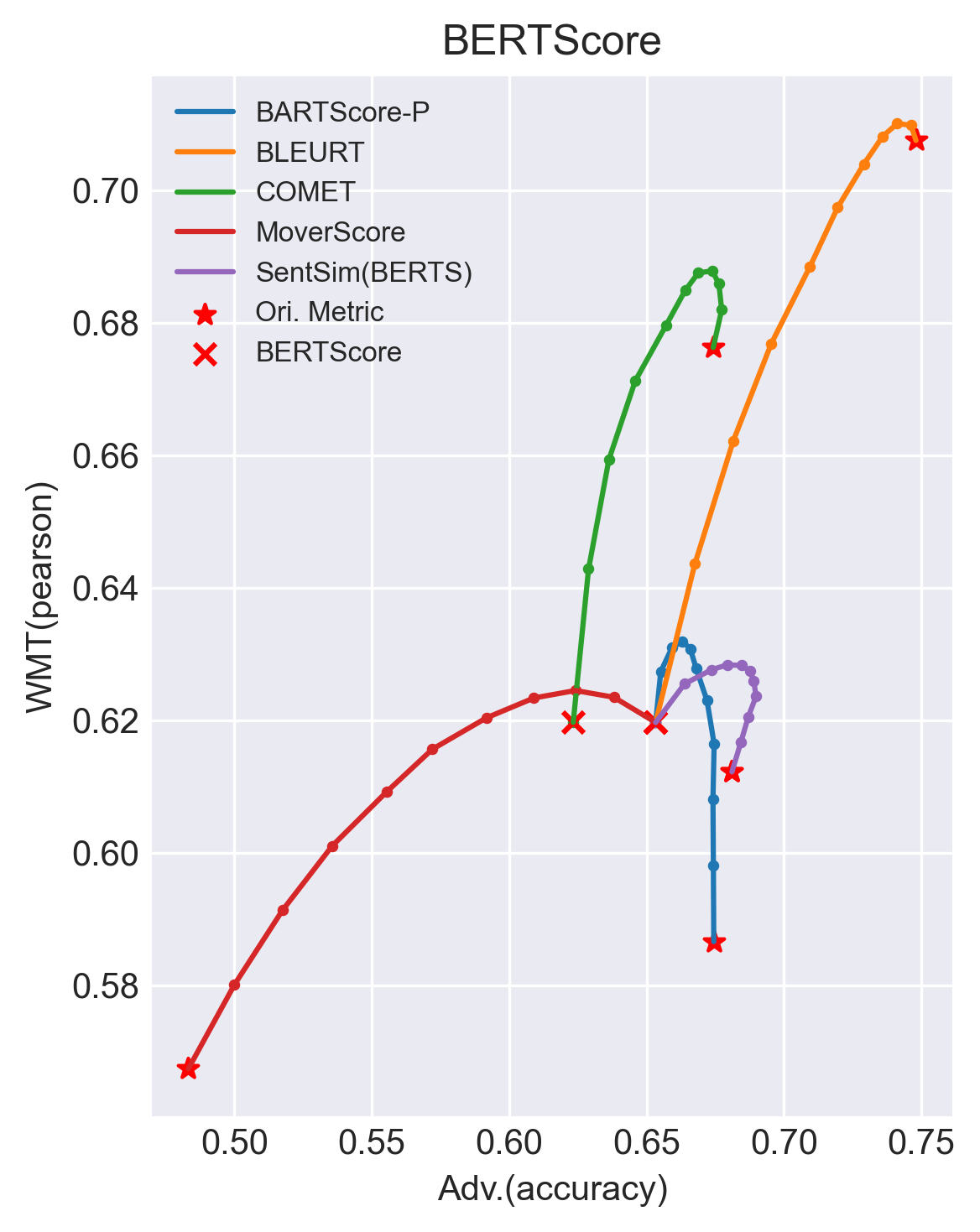}
    \caption{Accuracy on adversarial datasets and Pearson correlation with 
    {\sefinal{segment-level}} human judgements in WMT datasets of combined metrics with BERTScore, averaged over datasets. 
    The green line denoting the combination with COMET ends at another point since the corresponding adversarial performance is only averaged over the 2 adversarial datasets containing source texts.}
    \label{fig:ens_bertscore}
    \vspace{-.5cm}
\end{figure}
\ycrev{
We compare the ensembles with NLI metrics to ensembles with standard metrics, i.e., $w\cdot A+(1-w)\cdot M$ where $A$ is a fixed standard metric and $M$ 
is any of the remaining metrics. 
To do so, we combine 
standard metrics with the rest metrics for each category of MT/summarization and ref-based/-free setting. 
We 
take 
the arithmetic average of the accuracy on adversarial benchmarks and correlations
on standard benchmarks 
as the overall metric performance here. 
We calculate the mean/maximal improvement of ensembles to the original metric $M$ over $w \in [0.1,0.9]$ and observe: 
(i) While the ensembles with standard metrics are better for ref-free MT metrics 
because cross-lingual NLI metrics 
perform very poorly in our experiments,  
(ii) the monolingual NLI metrics 
lead to much better ensembles---17/15 points larger mean/max improvement---compared to the standard metrics. 
(iii) Overall, the ensembles with NLI metrics yield 
10/7 points larger mean/max improvement in 
overall performance than with standard metrics (averaged over all 4 tasks: ref-based/-free MT/summarization). 
\serev{
Thus, (monolingual) NLI metrics have unique properties, compared to 
standard metrics, making them attractive in ensembles.}%
}

\serev{To illustrate, Figure \ref{fig:ens_bertscore} shows ensembles with BERTScore. These show minor or no improvements on standard benchmarks and also mixed (often negative) results for adversarial robustness.}

\paragraph{SummaCZS and Falsesum:} 
\serev{In \S\ref{sec:results}, we applied NLI systems 
on whole input texts, not taking into account the multi-sentence nature of source texts and outputs, especially in summarization.}

To remedy the mismatch between the granularities of the training data of NLI models and the input data of summarization evaluation, i.e., sentence- vs.\  document-level,  \citet{laban-etal-2022-summac} propose both supervised and unsupervised NLI-based summarization 
metrics for inconsistency detection. 
We test their unsupervised variant (\textbf{SummaCZS}),\footnote{\yccr{We do not compare to the supervised one as it is trained on a consistency dataset for summarization task, for a fairer comparison.}} 
which segments documents into sentence units and aggregates scores between pairs of sentences, 
with the underlying model of \texttt{NLI-R}. 
However, 
SummaCZS does not consistently 
outperform \texttt{NLI-R} across all datasets; in contrast, \texttt{NLI-R} performs much better in our adversarial test compared to SummaCZS (72\% vs.\ 53\%). 
Besides, to match the training data of NLI models with the task of factual inconsistency detection in summarization,  \citet{https://doi.org/10.48550/arxiv.2205.06009} introduce an 
augmented 
NLI dataset with task-oriented examples based on CNNDM---\textbf{FalseSum}; we evaluate three Roberta-large models finetuned on it and MNLI. Similar to SummaCZS, this also does not always yield better performance compared to simple NLI metrics ($\sim$55\%–68\% vs.\ 72\% on adversarial datasets). Overall, both approaches work
well on SummEval, but not so well on RealSumm and 
our adversarial benchmark.

\paragraph{Choice of Pooling Strategy:}

\ycrev{
To examine the issue of data leakage discussed in \S\ref{sec:results}, 
we now evaluate the NLI metrics on each dataset with the pooling strategy selected from the remaining datasets (excluding the one for evaluation) based on winning frequency. For example, for the segment-level MT evaluation on WMT15, we choose the pooling strategy which wins most times on all MT datasets (including all standard datasets for both segment/system-level evaluation and the adversarial datasets) except for WMT15. 
We observe that this change in pooling strategy induction results in 
minor performance variation: -1.9\% for segment-level evaluation, +0.8\% for system-level evaluation and -0.7\% for adversarial evaluation. 
For summarization, 
as only one direction---i.e., \emph{src}$\rightarrow$\emph{cand}---is considered for ref-free NLI metrics, we separately select the pooling strategy for ref-based and ref-free NLI metrics. 
\serev{Overall, we have no performance change for the ref-free setting and 
\text{\sefinal{-3.6\%}} performance on average over all five criteria (correlations on SummEval with max/mean aggregation, summary/system-level correlations on RealSumm, and accuracy on SE$_{\text{adv}}$) ref-based. Thus, the changes are again minor.}%
}

\paragraph{Comparison to RoMe:} 
As the authors of RoMe did not publish their adversarial dataset, we compare RoMe's performance with our metrics on one of our adversarial datasets, 
WMT20$_\text{de}$, instead. 
RoMe has an average accuracy of 43\%, with > 90\% accuracy only on the phenomena \emph{SVD} and \emph{omission}, which are the easiest 
for most standard metrics. In contrast, our NLI metrics have above 80\% average accuracy. 
As RoMe does not evaluate on MT or summarization, we also evaluate our NLI metrics on one (randomly chosen) data-to-text generation dataset used in \citet{rony-etal-2022-rome}---BAGEL \citep{mairesse-etal-2010-phrase}. 
RoMe and our NLI metrics perform on par here ($\sim$0.23 Spearman's $\rho$).  
Overall, 
this seems to imply that simple NLI models taken out of the box are better and more robust metrics than a specially trained approach such as RoMe. 

\section{Concluding Remarks}\label{sec:conclusion}

In this work, we explored NLI as a \emph{general} paradigm for evaluation metrics. We showed that NLI metrics yield adversarial robustness, 
and are also strong---though not always state-of-the-art---when it comes to standard metric evaluation benchmarks. By linearly interpolating established (BERT-based) metrics with our NLI metrics, we obtained high-quality metrics along both axes: adversarial robustness and standard benchmarks, 
with 
substantial gains 
over 
recent BERT-based metrics.  

A potential reason why NLI based metrics perform subpar on some standard benchmarks (especially in MT) is the training data mismatch, i.e., typical NLI datasets contain many artificial sentences of the type ``A girl is playing on a piano''. 
A further limitation is that cross-lingual NLI models are not yet high-quality enough and that 
most current NLI models are sentence-level, not document-level---with a few recent exceptions \citep{yin-etal-2021-docnli}. Once these limitations of NLI 
are overcome, we believe that even better performances from NLI based metrics can be expected, which, we believe, is one of the most promising directions for future high-quality and robust evaluation metric \sefinal{design}. \serev{Future work should also consider NLI metrics for other text generation tasks; the NLI paradigm looks especially promising for tasks that require comparison with human references, which oftentimes involve the concept of logical equivalence.}

\section*{Acknowledgments}
We thank Zuojun Shi for conducting initial experiments related to this paper as part of her Bachelor thesis at TU Darmstadt. We appreciate the reviewers and editors from TACL for their time, effort, and greatly helpful comments. We also 
thankfully acknowledge support from the BMBF via the grant ``Metrics4NLG''. 
Steffen Eger is financed by DFG grant EG 375/5–1. 

\bibliography{custom}
\bibliographystyle{acl_natbib}

\end{document}